\documentclass[10pt,journal,compsoc]{IEEEtran}
\ifCLASSOPTIONcompsoc
  \usepackage[nocompress]{cite}
\else
  \usepackage{cite}
\fi

\usepackage{bm}
\usepackage{url}
\usepackage{amsfonts}
\usepackage{amsmath}
\usepackage{amssymb}
\usepackage{pifont}
\usepackage[caption=false]{subfig}
\usepackage{diagbox}
\usepackage{multirow}

\usepackage[align=center,shadow=true,shadowsize=4.7pt,nobreak=true,framemethod=tikz,skipabove=9.5pt,skipbelow=9pt,innertopmargin=5pt,innerbottommargin=5pt,innerleftmargin=5pt,innerrightmargin=5pt,leftmargin=1.5pt,rightmargin=1.5pt]{mdframed}
\usetikzlibrary{shadows}

\usepackage{color}
\usepackage{colortbl}
\definecolor{Gray}{gray}{0.93}
\definecolor{modify}{rgb}{0,0,0}

\usepackage[pagebackref=false,breaklinks=true,colorlinks,bookmarks=false]{hyperref}
\hypersetup{linkcolor=[rgb]{0.8,0.1,0.1}}
\hypersetup{citecolor=[rgb]{0.3,0.1,0.9}}

\newcommand{\cmark}{\ding{51}}
\newcommand{\xmark}{\ding{55}}

\newlength\savewidth\newcommand\shline{\noalign{\global\savewidth\arrayrulewidth
  \global\arrayrulewidth 1pt}\hline\noalign{\global\arrayrulewidth\savewidth}}

\newcommand{\eg}{\emph{e.g.}}
\newcommand{\ie}{\emph{i.e.}}
\newcommand{\norm}[1]{\left\lVert#1\right\rVert}
\begin{document}

\title{SphereFace Revived:\\Unifying Hyperspherical Face Recognition}

\author{Weiyang~Liu*,
        Yandong~Wen*,
        Bhiksha~Raj,~\IEEEmembership{Fellow,~IEEE},
        Rita Singh,
        Adrian~Weller
\IEEEcompsocitemizethanks{\IEEEcompsocthanksitem W. Liu and A. Weller are with the Department of Engineering, University of Cambridge, United Kingdom. W. Liu is also with the Max Planck Institute for Intelligent Systems, Tübingen, Germany. A. Weller is also with The Alan Turing Institute, London, United Kingdom.\protect\\
E-mail: wl396@cam.ac.uk, aw665@cam.ac.uk
\IEEEcompsocthanksitem Y. Wen, B. Raj and R. Singh are with the Department of Electrical and Computer Engineering, Carnegie Mellon University, United States.\protect\\
E-mail: yandongw@andrew.cmu.edu, \{bhiksha,rsingh\}@cs.cmu.edu
\IEEEcompsocthanksitem *W. Liu and Y. Wen have contributed equally to this work.
}
\thanks{Accepted for publication on 1 March 2022}
}

\markboth{IEEE TRANSACTIONS ON PATTERN ANALYSIS AND MACHINE INTELLIGENCE}
{Shell \MakeLowercase{\textit{et al.}}: Bare Demo of IEEEtran.cls for Computer Society Journals}

\IEEEtitleabstractindextext{%
\begin{abstract}
This paper addresses the deep face recognition problem under an open-set protocol, where ideal face features are expected to have smaller maximal intra-class distance than minimal inter-class distance under a suitably chosen metric space. To this end, hyperspherical face recognition, as a promising line of research, has attracted increasing attention and gradually become a major focus in face recognition research. As one of the earliest works in hyperspherical face recognition, SphereFace explicitly proposed to learn face embeddings with large inter-class angular margin. However, SphereFace still suffers from severe training instability which limits its application in practice. In order to address this problem, we introduce a unified framework to understand large angular margin in hyperspherical face recognition. Under this framework, we extend the study of SphereFace and propose an improved variant with substantially better training stability -- SphereFace-R. Specifically, we propose two novel ways to implement the multiplicative margin, and study SphereFace-R under three different feature normalization schemes (no feature normalization, hard feature normalization and soft feature normalization). We also propose an implementation strategy -- ``characteristic gradient detachment'' -- to stabilize training. Extensive experiments on SphereFace-R show that it is consistently better than or competitive with state-of-the-art methods.

\end{abstract}

\begin{IEEEkeywords}
Hypersphere, face recognition, angular margin, loss function
\end{IEEEkeywords}}

\maketitle

\IEEEdisplaynontitleabstractindextext

\IEEEpeerreviewmaketitle

\IEEEraisesectionheading{\section{Introduction}\label{sec:introduction}}

\IEEEPARstart{R}{ecent} years have witnessed the tremendous success of deep face recognition~(FR). Owing to the rapid development in discriminative loss functions~\cite{liu2016large,liu2017sphereface,wang2018additive,wang2018cosface,deng2019arcface} that promote large inter-class feature margin, the performance of deep FR has dramatically improved. 
These loss functions share a common goal to project deeply learned face embeddings onto a hypersphere and incorporate large geodesic inter-class margins. We call this series of deep FR methods \emph{hyperspherical face recognition}.

Previous deep FR methods~\cite{taigman2014deepface,sun2014deep} typically train neural networks by classifying identities in a training set. Such a training target largely deviates from the open-set testing (\ie, to determine whether two face images belong to the same person) in two aspects: \emph{(i)} similarity measure differs in training and testing; \emph{(ii)} open-set testing must solve a metric learning problem~\cite{liu2017sphereface} where the goal is to learn large-margin features, while training aims to solve a closed-set classification problem where the goal is to learn separable features. Motivated by the mismatch between training and testing in deep FR, hyperspherical FR aims to bridge the gap by \emph{(i)} constraining the face embeddings on the hypersphere (\ie, using cosine similarity for both training and testing), and \emph{(ii)} incorporating large geodesic margin on the hypersphere. Another motivation for hyperspherical FR comes from the observation that deep features are intrinsically discriminative on a hypersphere~\cite{liu2018decoupled}. Hyperspherical FR essentially focuses on answering the question: \emph{How to effectively and stably incorporate large angular margin to face embeddings?}

Large-margin softmax~\cite{liu2016large} is one of the first methods to incorporate large angular margin to deeply learned features. The core idea is to use a monotonically decreasing lower bound function $\psi(\theta_{(\bm{x},\bm{W}_y)})$ to replace the target angular activation $\cos(\theta_{(\bm{x},\bm{W}_y)})$ in the softmax-based loss function, where $\theta_{(\bm{x},\bm{W}_y)}$ denotes the angle between deep feature $\bm{x}$ and the classifier of the target class $\bm{W}_y$ ($y$ is the label of $\bm{x}$). The intuition is that the function $\psi(\theta_{(\bm{x},\bm{W}_y)})$ will make the angle $\theta_{(\bm{x},\bm{W}_y)}$ smaller in order to achieve the same value of $\cos(\theta_{(\bm{x},\bm{W}_y)})$. Such a design will encourage the deep features to have large inter-class margins on the unit hypersphere. Most popular hyperspherical FR methods~\cite{liu2017sphereface,wang2018additive,wang2018cosface,deng2019arcface} adopt this design principle.

Built upon \cite{liu2016large}, SphereFace~\cite{liu2017sphereface} takes one step further by explicitly constraining decision boundaries on the hypersphere and simultaneously incorporating angular margins. Inspired by SphereFace, there is a series of work~\cite{wang2018additive,wang2018cosface,deng2019arcface} that design alternative lower bound target function $\psi(\theta_{(\bm{x},\bm{W}_y)})$ to achieve angular margin. Based on how $\psi(\theta_{(\bm{x},\bm{W}_y)})$ is constructed, loss functions in hyperspherical FR can be divided into \emph{additive margin}~\cite{wang2018additive,wang2018cosface,deng2019arcface} and \emph{multiplicative margin}~\cite{liu2016large,liu2017sphereface}. As a representative  multiplicative margin method, SphereFace renders promising geometric insights. However, in contrast to additive margin, SphereFace is known to be highly non-trivial to train, typically requiring a number of bells and whistles to stabilize its training, which limits its potential application.

In order to address this shortcoming, we take a detour by first identifying an intrinsic connection that bridges different margin designs~\cite{liu2016large,liu2017sphereface,wang2017normface,wang2018additive,wang2018cosface,deng2019arcface} in hyperspherical FR. We formulate this connection with a unified large-margin framework for hyperspherical FR. In this framework, we summarize a general principle for any loss function to achieve large angular margin. Following this principle, most existing hyperspherical FR methods can be viewed as special instantiations. This framework 
helps us gain a deeper understanding 
of hyperspherical FR, and serves as a portal to design new loss functions.

Under this unified framework, we extend our previous study of SphereFace~\cite{liu2017sphereface} by proposing alternative yet effective ways to implement the multiplicative margin with improved training stability and better empirical performance. Specifically, the original realization of multiplicative margin in SphereFace is exact only when the angle between the feature and the target classifier is sufficiently small. When this angle is large, the multiplicative margin becomes approximate and the original intuition no longer holds. Motivated by this, we propose two novel variants that can exactly implement the intuition of multiplicative margin for all possible angles. Along with the new multiplicative margins, we also propose a novel implementation strategy which we call \emph{characteristic gradient detachment}~(CGD) that helps to stabilize training and improve generalization. We term our improved approach \emph{SphereFace-R}.

Another significant difference between SphereFace and other hyperspherical FR methods~\cite{wang2017normface,wang2018additive,wang2018cosface,deng2019arcface} is whether feature normalization~(FN) is performed. Based on the empirical observation in \cite{liu2017sphereface,liu2018decoupled}, we notice that feature magnitude still contains some information such as image quality. However, whether the information encoded in feature magnitude is useful for FR remains an open question. To address this, we consider three schemes here: no feature normalization~(NFN), hard feature normalization~(HFN) and soft feature normalization~(SFN). HFN is identical to the popular feature normalization used in \cite{wang2017normface,wang2018additive,wang2018cosface,deng2019arcface}. In contrast, SFN formulates the feature normalization objective into a regularization term and optimizes it jointly with the neural network. Unlike HFN, SFN will take feature magnitude into account when training the neural network. This shares a similar spirit with \cite{zheng2018ring}. While both FN-free learning and HFN can be viewed as limiting cases of SFN, SFN effectively unifies both approaches and serves as an interpolation between them. We conduct a systematic study to evaluate the effectiveness of all three FN strategies.

\vspace{1.5mm}
Our contributions can be summarized as follows:
\vspace{-1.5mm}

\begin{itemize}
    \item We present a unified framework to understand large angular margin in hyperspherical FR. This framework effectively explains how and why angular margin can be incorporated in SphereFace and further summarizes a general principle for loss functions to introduce large angular margin. Moreover, most of the current hyperspherical FR methods can be viewed as special instantiations of this framework.
    \item Under the unified framework, we substantially extend our previous work on SphereFace~\cite{liu2017sphereface} by addressing training instability and improving empirical performance. Compared to the original SphereFace, SphereFace-R uses a more intuitive way to incorporate the multiplicative margin and yields more stable training, more clear geometric interpretation and superior generalization.
    \item We propose CGD, a generic implementation method for hyperspherical FR methods to improve the training stability and generalizability.
    \item To evaluate the usefulness of feature magnitude, we comprehensively study SphereFace-R under three different FN schemes: NFN, HFN and SFN. 
    \item Our paper comes with an easy-to-use codebase to facilitate future research.\footnote{See \textbf{Project OpenSphere}: \url{https://opensphere.world/}.} It serves as a platform to evaluate hyperspherical FR methods fairly.
\end{itemize}


\section{Related Work}

\textbf{Deep face recognition}. Deep face recognition 
has been an active research area in the past decades. \cite{taigman2014deepface,sun2014deep,parkhi2015deep} address open-set FR using a convolutional neural network~(CNN) supervised by softmax-based loss, which essentially views open-set FR as a multi-class classification problem. \cite{sun2014deep2} combines contrastive loss and softmax loss to jointly supervise the CNN training, greatly boosting performance. \cite{schroff2015facenet} uses triplet loss and feature normalization to learn a unified face embedding. After training on nearly 200 million face images, they achieve state-of-the-art performance. Inspired by linear discriminant analysis, \cite{wen2016discriminative} propose center loss for CNNs and 
obtain promising performance. Prior to hyperspherical FR, well-performing deep FR methods~\cite{sun2016sparsifying,liu2015targeting,schroff2015facenet} were mostly built on either contrastive loss or triplet loss, validating the importance of solving open-set FR as a metric learning problem~\cite{liu2017sphereface}. The development of deep FR is also closely related to deep metric learning~\cite{hu2014discriminative,lu2015multi,song2016deep,movshovitz2017no,wu2017sampling,ge2018deep,duan2018deep,qian2019softtriple,wang2019ranked,sun2020circle,musgrave2020metric,wen2019comprehensive}.

\vspace{0.7mm}
\noindent\textbf{Hyperspherical face recognition}. As a major line of research in deep FR, hyperspherical FR~\cite{liu2017sphereface,ranjan2017l2,liu2017rethinking,liu2017deep,wang2017normface,wang2018additive,wang2018cosface,zheng2018ring,liu2018learning,ranjan2018crystal,deng2019arcface,zhao2019regularface,duan2019uniformface,zhang2019adacos,wen2019comprehensive,liu2019adaptiveface,wu2020rotation,kim2020discface,sun2020circle,huang2020curricularface,deng2020sub,kim2020groupface,zhong2021sface,li2021spherical,meng2021magface,wen2022sphereface2} has become increasingly popular in recent years due to its effectiveness. \cite{liu2016large} proposes the initial framework of learning deep features with large angular margin. Built upon this framework, SphereFace~\cite{liu2017sphereface} normalizes the classifier weights and explicitly models the decision boundary on the hypersphere. Feature normalization~\cite{ranjan2017l2,liu2017rethinking,wang2017normface} has also been introduced to facilitate the learning of large angular margin face embeddings. To improve training stability, CosFace~\cite{wang2018additive,wang2018cosface} and ArcFace~\cite{deng2019arcface} propose to use additive angular margin to replace the original multiplicative margin in SphereFace and obtain impressive performance. \cite{zhang2019adacos} and \cite{liu2019adaptiveface} consider adaptive schemes to set the radius of the hypersphere and the margin parameter, respectively. \cite{sun2020circle,huang2020curricularface} study hyperspherical FR by taking the easy-hard sample balance into consideration. 

\vspace{0.7mm}
\noindent\textbf{Hyperspherical learning}. Beyond face recognition, the idea of learning a representation on the hypersphere is also shown generally useful in a diverse set of applications, such as few-shot recognition~\cite{chen2019closer,liu2019neural,mettes2019hyperspherical,liu2021learning,liu2021orthogonal}, deep metric learning~\cite{wang2019ranked,yu2019deep}, self-supervised learning~\cite{he2020momentum,chen2020simple,grill2020bootstrap,wang2020understanding}, generative models~\cite{park2019sphere,davidson2018hyperspherical}, geometric learning~\cite{cohen2018spherical,rao2019spherical,lin2020regularizing}, person re-identification~\cite{fan2019spherereid,yu2019unsupervised,hao2019hsme,sun2020circle}, speech processing~\cite{hajibabaei2018unified,liu2019large,li2019boundary,fathullah2020improved} and text processing~\cite{meng2019spherical}. It has been widely observed that constraining the embedding space on a hypersphere is beneficial to generalizability.

In contrast to hyperspherical FR methods that use an additive margin, SphereFace-R is built upon our previous work~\cite{liu2016large,liu2017sphereface} and adopts a multiplicative margin approach. 

\begin{table*}[t]
	\centering
	\setlength{\tabcolsep}{2.5pt}
	\renewcommand{\arraystretch}{1.5}
	\caption{Instantiations of the Unified Hyperspherical Face Recognition Framework. Gray region denotes our contribution.}
	\vspace{-2mm}
	\begin{tabular}{l|c|c|c|c}
		Method & Feature Magnitude  & Non-target Function $\eta(\theta)$ & Target Function $\psi(\theta)$ & $\Delta(\theta)$ (shown in Fig.~\ref{fig:delta_func})\\
		\shline
		NormFace~\cite{wang2017normface} & HFN  & $\cos(\theta)$ & $\cos(\theta)$ & $0$ \\
		CosFace~\cite{wang2018additive,wang2018cosface} &  HFN & $\cos(\theta)$ & $\cos(\theta)-m$ & $m$ \\
		ArcFace~\cite{deng2019arcface} & HFN & $\cos(\theta)$ & $\cos(\theta+m)$ & $\cos(\theta)-\cos(\theta+m)$ \\
		\hline
		\rowcolor{Gray}
		SphereFace & NFN~\cite{liu2017sphereface}/HFN/SFN & $\cos(\theta)$ & $(-1)^k\cos(m\theta)-2k$, $\theta\in[\frac{k\pi}{m}, \frac{(k+1)\pi}{m} ],k\in\mathbb{N}$ & $\cos(\theta)-(-1)^k\cos(m\theta)+2k$\\
		\rowcolor{Gray}
		SphereFace-R v1 & NFN/HFN/SFN & $\cos(\theta)$ & $\cos(\min\{ m,\frac{\pi}{\theta} \}\cdot\theta)$ & $\cos(\theta)-\cos(\min\{ m,\frac{\pi}{\theta} \}\cdot\theta)$\\
		\rowcolor{Gray}
		SphereFace-R v2 & NFN/HFN/SFN & $\cos(\frac{\theta}{m})$ & $\cos(\theta)$ & $\cos(\frac{\theta}{m})-\cos(\theta)$\\
	\end{tabular}
	\label{framework_case}
\end{table*}

\section{A Unified Large-Margin Learning Framework for Hyperspherical Face Recognition}\label{unified_framework}

To gain deeper insights towards large angular margin, we present a unified framework for hyperspherical FR. To start with, we consider the standard softmax cross-entropy loss:
\begin{equation}
    \mathcal{L}_s= -\log\bigg(\frac{\exp(\bm{W}_y^\top\bm{x}+b_y)}{\sum_{i=1}^K \exp(\bm{W}_i^\top\bm{x}+b_i)}\bigg)
\end{equation}
where $\bm{x}\in\mathbb{R}^d$ denotes the deep feature (the input of the classifier layer), $y$ is its ground truth label, $K$ is the total number of classes, $\bm{W}_i\in\mathbb{R}^d$ is the weights of the $i$-th classifier and $b_i$ is the bias for the $i$-th class. Note that here we consider the case of a single input sample for simplicity; we only need to average the loss objectives if we consider a mini-batch of input samples. Since the class-dependent bias term is not informative in open-set evaluation, we follow the common practice to remove it~\cite{liu2017sphereface}. Then we normalize the classifier weights to one (\ie, $\|\bm{W}_i\|=1,\forall i$) and rewrite the objective function as follows:
\begin{equation}
    \mathcal{L}_s= -\log\bigg(\frac{\exp\big(\norm{\bm{x}}\cos(\theta_y)\big)}{\sum_{i=1}^K \exp(\norm{\bm{x}}\cos(\theta_i))}\bigg)
\end{equation}
where $\theta_i$ denotes the angle between deep feature $\bm{x}$ and the $i$-th classifier $\bm{W}_i$. By considering a generic angular activation rather than the cosine function, we have the following generalized objective function:
\begin{equation}\label{generalized-softmax}
    \mathcal{L}_g=\! -\log\bigg(\frac{\exp\big(\norm{\bm{x}}\psi(\theta_y)\big)}{\exp\big(\norm{\bm{x}}\psi(\theta_y)\big)\!+\!\sum_{i\neq y} \exp(\norm{\bm{x}}\eta(\theta_i))}\bigg)
\end{equation}
where $\psi(\theta_y)$ is the angular activation function for the target class (\ie, ground truth label) and $\eta(\theta_i),i\neq y$ denotes the angular activation function for the $i$-th non-target class (the labels excluding the ground truth one). Similar to the cosine function, both $\psi(\theta)$ and $\eta(\theta)$ are generally required to be monotonically decreasing for $\theta\in[0,\pi]$. After looking into different hyperspherical FR methods, we summarize a simple yet generic principle for any softmax loss in order to learn embeddings with large angular margin.
\begin{mdframed}
To achieve large angular margin, the generic principle is to make $\psi(\theta)$ always smaller than $\eta(\theta)$ in $(0, \pi]$, namely
\begin{equation}\label{margin-principle}
    \Delta(\theta)=\eta(\theta)-\psi(\theta)>0
\end{equation}
where we define $\Delta(\theta)$ as the characteristic function for large angular margin. $\Delta(\theta)$ determines most of the properties about the angular margin, such as the size of the margin, its learning stability, etc.
\end{mdframed}

As long as we guarantee that $\Delta(\theta)$ is larger than zero, then the objective function in Eq.~\eqref{generalized-softmax} will define a task that can inherently introduce large angular margin. To see how $\Delta(\theta)$ interacts with the loss function, we can rewrite Eq.~\eqref{generalized-softmax} in the following mathematically equivalent form:
\begin{equation}\label{new-softmax}
\begin{aligned}
    \mathcal{L}_g &= \log \Big( 1 + \sum_{i\neq y}\exp\big(\norm{\bm{x}}(\eta(\theta_i)-\psi(\theta_y))\big) \Big)\\
    &=\log \Big( 1 + \sum_{i\neq y}\exp\big(\norm{\bm{x}}(\eta(\theta_i)-\eta(\theta_y)+\Delta(\theta_y))\big) \Big)
\end{aligned}
\end{equation}
which essentially aims to minimize $\eta(\theta_i)-\eta(\theta_y)+\Delta(\theta_y)$. The term $\eta(\theta_i)-\eta(\theta_y)$ represents the difference of classification confidence, and the characteristic function $\Delta(\theta_y)$ controls the angular margin. When $\Delta(\theta_y)=0$ and $\eta(\cdot)$ is the cosine function, Eq.~\eqref{new-softmax} reduces to the standard softmax loss with weight normalization. $\Delta(\theta_y)=0$ indicates that no angular margin has been introduced. When $\Delta(\theta_y)>0$, this leads to large angular margin because it makes the classification more stringent (\ie, the neural network will learn to make $\theta_y$ smaller in order to reach the same loss value as the case of $\Delta(\theta_y)=0$). It is also worth mentioning that when $\Delta(\theta_y)<0$, Eq.~\eqref{new-softmax} defines an easier task than the standard classification problem and is potentially useful for robust learning against noisy images or labels. Our paper focuses on the case of $\Delta(\theta_y)>0$.

We note that there exist scenarios where large angular margin can still be achieved even if $\Delta(\theta_y)$ is smaller than zero in some range of $\theta_y\in[0,\theta]$. For example, $\Delta(\theta_y)$ for ArcFace can be smaller than zero when $\theta_y$ is close to $\pi$. ArcFace can still introduce angular margin because the case where $\theta_y$ is close to $\pi$ hardly happens with real data distribution, as verified by \cite{deng2019arcface}. Nonetheless, the characteristic function for ArcFace still approximately satisfies our principle, since it is larger than zero with most $\theta_y\in[0,\theta]$. Therefore, as long as the characteristic function $\Delta(\theta_y)$ is larger than zero for the angles where $\theta_y$ is densely distributed in practice (\ie, $\mathbb{E}_{\theta_y}\Delta(\theta_y)>0$), it will typically suffice to produce effective angular margin. Our principle in fact serves as a sufficient condition to introduce angular margin. It is generally better to use our principle as the guideline for designing new angular margin losses, because the empirical distribution of the target angle $\theta_y$ could vary under difference circumstances (\eg, network architectures, datasets, optimizers).

We now discuss in depth why $\Delta(\theta_y)>0$ is able to introduce large angular margin. For ease of illustration, we consider the binary case where the first class is the target class. In this case, we only need to discuss $\eta(\theta_2)-\eta(\theta_1)+\Delta(\theta_1)$. If $\Delta(\theta_1)=0$, the decision boundary for the first class is $\eta(\theta_2)-\eta(\theta_1)=0$ which is equivalent to $\theta_1=\theta_2$. When $\theta_1<\theta_2$, the sample $\bm{x}$ will be classified to the first class. If $\Delta(\theta_1)>0$ and $\eta(\cdot)$ is monotonically decreasing, then the decision boundary for the first class becomes $\eta(\theta_2)-\eta(\theta_1)+\Delta(\theta_1)=0$ which is equivalent to $\theta_1+m(\theta_1)=\theta_2$ where $m(\cdot)$ denotes some positive function (the specific form of $m(\cdot)$ is determined by $\eta(\cdot)$ and $\Delta(\cdot)$, but it stays positive as long as $\Delta(\cdot)$ is always positive). Therefore, now we need to make $\theta_1+m(\theta_1)<\theta_2$ in order to classify $\bm{x}$ to the first class, and the decision boundary for the first class becomes more stringent than the previous case. The neural network has to learn smaller $\theta_1$ in order to correctly classify $\bm{x}$ and smaller $\theta_1$ implies a more compact representation for the first class. The same reasoning also applies to the case where $\bm{x}$ belongs to the second class (\ie, the second class is the target class). As a result, if we can successfully train a neural network to correctly classify training samples with these more stringent classification criteria, $\Delta(\theta_1)>0$ can effectively produce large angular margin for the learned deep features.

\begin{figure}[t]
  \centering
 \includegraphics[width=3.535in]{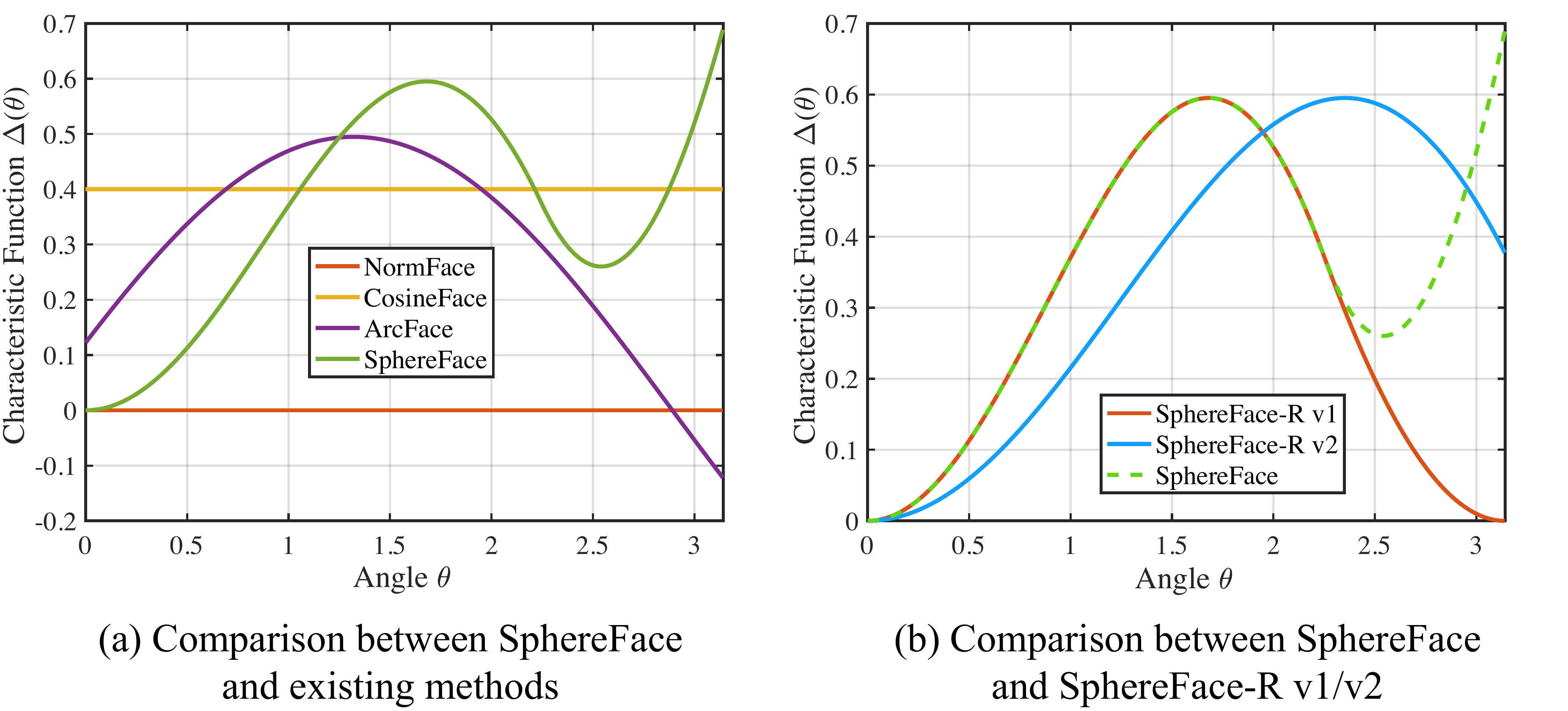}
  \vspace{-6.4mm}
  \caption{(a) $\Delta(\theta)$ for current representative hyperspherical FR methods. (b) $\Delta(\theta)$ of SphereFace and SphereFace-R. We set $m$ to $0.4$, $0.5$, $1.4$, $1.4$ and $1.4$ for CosFace, ArcFace, SphereFace, SphereFace-R v1 and SphereFace-R v2, respectively.}\label{fig:delta_func}
\end{figure}

Importantly, current popular hyperspherical FR methods can be viewed as special cases under this unified framework, as shown in Table~\ref{framework_case} (first four rows). To intuitively understand different variants of angular margin, we also compare their characteristic functions $\Delta(\theta)$ in Fig.~\ref{fig:delta_func}(a). One can observe that different hyperspherical FR methods yield different large-margin characteristic functions. Each characteristic function determines how a hyperspherical FR method performs and therefore it is of great significance to design a suitable characteristic function. Specifically, the characteristic function $\Delta(\theta)$ clearly reveals the induced angular margin for samples with different recognition hardness (larger $\theta_y$ typically implies a harder sample). Instead of a static characteristic function, designing a dynamic characteristic function could be beneficial~\cite{liu2019adaptiveface,huang2020curricularface}. It is also possible to learn the characteristic function in a data-driven and automatic fashion, as explored in \cite{wang2020loss,li2019lfs}.

Besides the characteristic function $\Delta(\cdot)$, the feature magnitude $\|\bm{x}\|$ in Eq.~\eqref{new-softmax} also plays a non-negligible role in learning large angular margin. The original SphereFace approach preserves the feature magnitude in training, since the feature magnitude does not affect the angular decision boundary. \cite{ranjan2017l2,wang2017normface,wang2018additive,wang2018cosface,deng2019arcface} show that normalizing the feature magnitude to a constant $s$ (\eg, making $\bm{x}\leftarrow s\frac{\bm{x}}{\|\bm{x}\|}$ in Eq.~\eqref{new-softmax}) can stabilize training and also improve hyperspherical discriminativeness. By normalizing the feature magnitude to a prescribed positive constant $s$, Eq.~\eqref{new-softmax} becomes
\begin{equation}\label{normalized-softmax}
    \mathcal{L}_{s} = \log \Big( 1 + \sum_{i\neq y}\exp\big(s\cdot(\eta(\theta_i)-\eta(\theta_y)+\Delta(\theta_y))\big) \Big)
\end{equation}
where $s$ is a universal value instead of the original instance-dependent $\|\bm{x}\|$. There are two advantages of feature normalization. First, it can effectively avoid potential bad local minima. Second, it can help the loss function to better balance easy and hard training samples. 

\begin{figure}[t]
  \centering
 \includegraphics[width=3.52in]{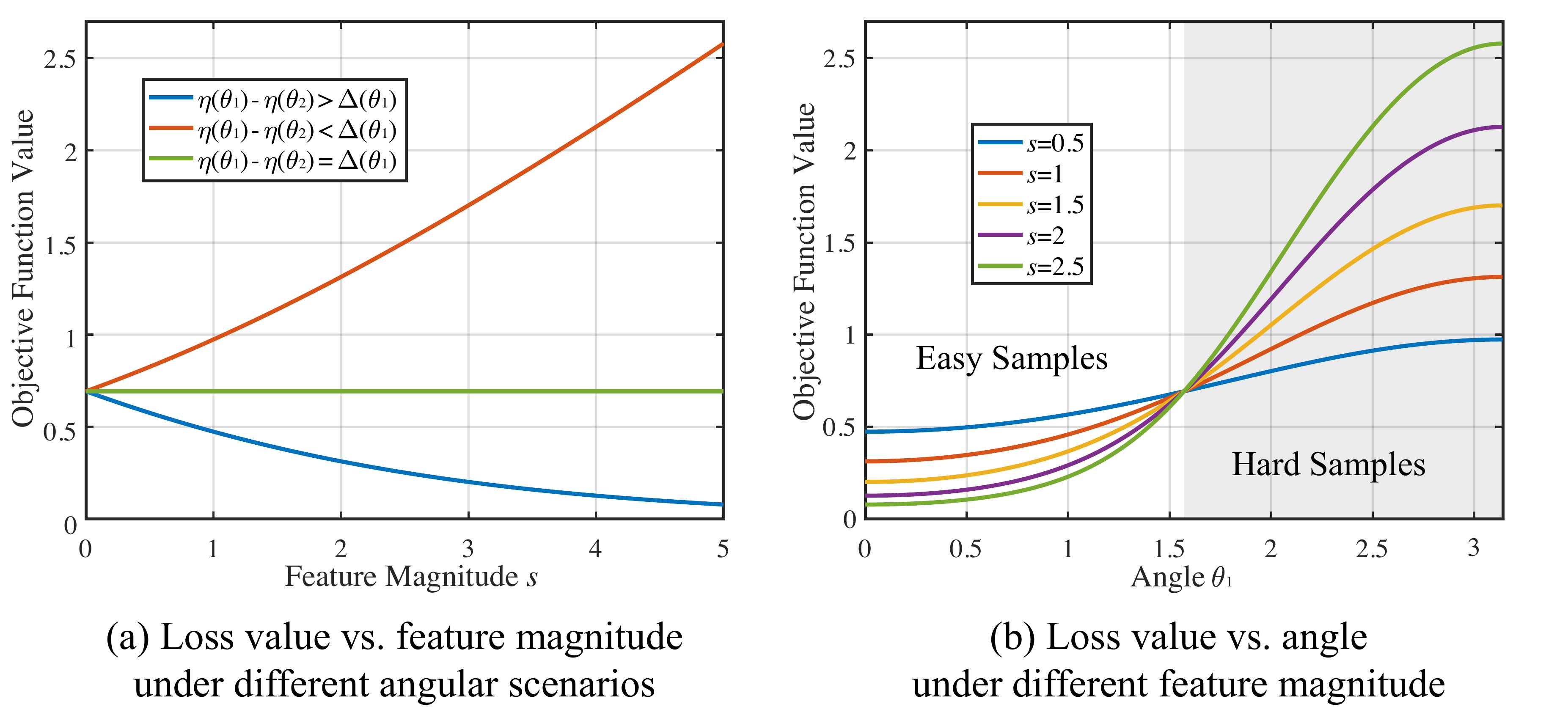}
  \vspace{-6.4mm}
  \caption{(a) How the loss value changes as feature magnitude $s$ increases under different angular scenarios in Eq.~\eqref{fn_limit}. For this figure, we consider the binary case where $\theta_1=\pi/3$ ($y=1$) and $\theta_2=\pi/2$. Both $\eta(\cdot)$ and $\psi(\cdot)$ are cosine function. (b) How the loss curve of Eq.~\eqref{normalized-softmax} varies under different feature magnitude $s$. For this figure, we consider the binary case where $y=1$ and $\theta_2=\pi/2$.}\label{fig:fn}
\end{figure}

For the first aspect, we consider Eq.~\eqref{normalized-softmax} in a simple binary classification scenario (class 1 is the ground truth label for the deep feature $\bm{x}$, \ie, $y=1$). The loss value can easily go to zero once the deep feature $\bm{x}$ lies in the correct decision region, as demonstrated in the following equation:
\begin{equation}\label{fn_limit}
\begin{aligned}
&\lim_{s\rightarrow +\infty} \log \Big( 1 + \exp\big(s\cdot(\eta(\theta_2)-\eta(\theta_1)+\Delta(\theta_1))\big) \Big)\\
=&\left\{
{\begin{array}{*{20}{l}}
0 & \textnormal{if} &\eta(\theta_1)-\eta(\theta_2)>\Delta(\theta_1)\\
-\log\frac{1}{2} & \textnormal{if} &\eta(\theta_1)-\eta(\theta_2)=\Delta(\theta_1)\\
+\infty & \textnormal{if} & \eta(\theta_1)-\eta(\theta_2)<\Delta(\theta_1)
\end{array}} \right.
\end{aligned}
\end{equation}
where $\eta(\theta_1)-\eta(\theta_2)>\Delta(\theta_1)$ means that $\bm{x}$ can be correctly classified, $\eta(\theta_1)-\eta(\theta_2)=\Delta(\theta_1)$ means that $\bm{x}$ exactly lies on the decision boundary and $\eta(\theta_1)-\eta(\theta_2)<\Delta(\theta_1)$ means that $\bm{x}$ can not be correctly classified. The results imply that when $\bm{x}$ can be correctly classified, a trivial solution to reduce loss to zero is to simply increase $s$. However, increasing $s$ does not help the neural network learn angularly discriminative face embeddings and results in bad local minima. Because $\|\bm{x}\|$ can be viewed as an instance-dependent learnable $s$, the neural network without feature normalization is likely to simply increase $\|\bm{x}\|$ after $\theta_1$ passes the decision boundary. Therefore, using a constant $s$ can prevent this trivial way of reducing loss value and eliminate these bad local minima. We also plot how the loss value changes as $s$ increases in Fig.~\ref{fig:fn}(a). The same argument can easily generalize to the multi-class scenario, as shown in
\begin{equation}
\begin{aligned}
&\lim_{s\rightarrow +\infty} \log \Big( 1 + \sum_{i\neq y}\exp\big(s(\eta(\theta_i)-\eta(\theta_y)+\Delta(\theta_y))\big) \Big)\\
=&\left\{
{\begin{array}{*{20}{l}}
0 & \textnormal{if} & \forall i\neq y,~\eta(\theta_y)-\eta(\theta_i)>\Delta(\theta_y)\\
+\infty & \textnormal{if} & \exists i\neq y,~\eta(\theta_y)-\eta(\theta_i)<\Delta(\theta_y)
\end{array}} \right.
\end{aligned}
\end{equation}
where $s$ has a large influence on the loss value. For the multi-class scenario, the neural network can trivially increase $s$ to minimize the loss once $\eta(\theta_y)-\eta(\theta_i)>\Delta(\theta_y)$ for all $i\neq y$. Interestingly, this also explains why the standard softmax loss cannot learn deep features with large angular margin. Empirically the standard softmax loss tends to increase $s$ instead of minimizing the target angle once the deep feature $\bm{x}$ falls into the correct decision boundary, leading to separable features rather than large-margin features. Large-margin losses take advantage of this phenomenon and make the decision boundary asymmetric for different classes (\ie, $\eta(\theta)\neq\psi(\theta)$). Then the classification of $\bm{x}$ (\ie, forcing $\psi(\theta_y)>\eta(\theta_i),\forall i\neq y$) naturally becomes equivalent to learning large-margin deep features.

For the second aspect, we use an example to demonstrate how the feature magnitude $s$ can balance the easy and hard samples. We compare the loss function under different $s$ in Fig.~\ref{fig:fn}(b). By adjusting $s$, the loss function in Eq.~\eqref{normalized-softmax} has different sensitivity for samples with different target angle $\theta_y$. Intuitively, samples with large target angle are considered to be hard, while samples with small target angle are viewed as easy. Therefore, feature magnitude $s$ can also balance the loss value for easy and hard samples, which serves a role similar to hard sample mining~\cite{wu2017sampling} in deep metric learning. From Fig.~\ref{fig:fn}(b), one can observe that larger $s$ puts more focus on the hard samples, since the loss ratio between hard and easy samples increases. 
Finding a good $s$ essentially can be viewed as searching for a suitable balance between easy and hard samples in hyperspherical FR methods, \eg, \cite{wang2017normface,wang2018additive,wang2018cosface,deng2019arcface}.

To summarize, Eq.~\eqref{normalized-softmax} essentially throws away the information encoded in the feature magnitude $\bm{x}$. Despite the two major advantages that ease the training of hyperspherical FR methods, it remains an open problem whether it is beneficial to combine feature magnitude to training. Feature magnitude is closely related to image quality and semantic ambiguity~\cite{liu2018decoupled,chen2020angular}, and such information intuitively seems useful to distinguish different faces. However, training hyperspherical FR methods without feature normalization generally yields inferior training stability and generalization performance in practice. In order to explore whether feature magnitude is indeed helpful or not, we consider to constrain the feature magnitude via a soft regularization in Section~\ref{section:mag}. This serves as an interpolation between no feature normalization and hard feature normalization, and can take the feature magnitude into account during training.

\section{SphereFace-R: Better and More Stable}

In this section, we elaborate the design of SphereFace-R and introduce two novel variants that perform well in practice. 
Sharing the same geometric interpretation as SphereFace, SphereFace-R yields improved training stability and superior open-set generalizability. We start by revisiting the design of the original SphereFace and then propose alternative ways to implement the multiplicative margin in Section~\ref{section:rethink}. In Section~\ref{section:mag}, we discuss different feature normalization schemes. Finally, we list a few important open problems for hyperspherical FR in Section~\ref{section:diss}.

\subsection{Rethinking Multiplicative Angular Margin}\label{section:rethink}

\begin{figure}[t]
  \centering
 \includegraphics[width=3.35in]{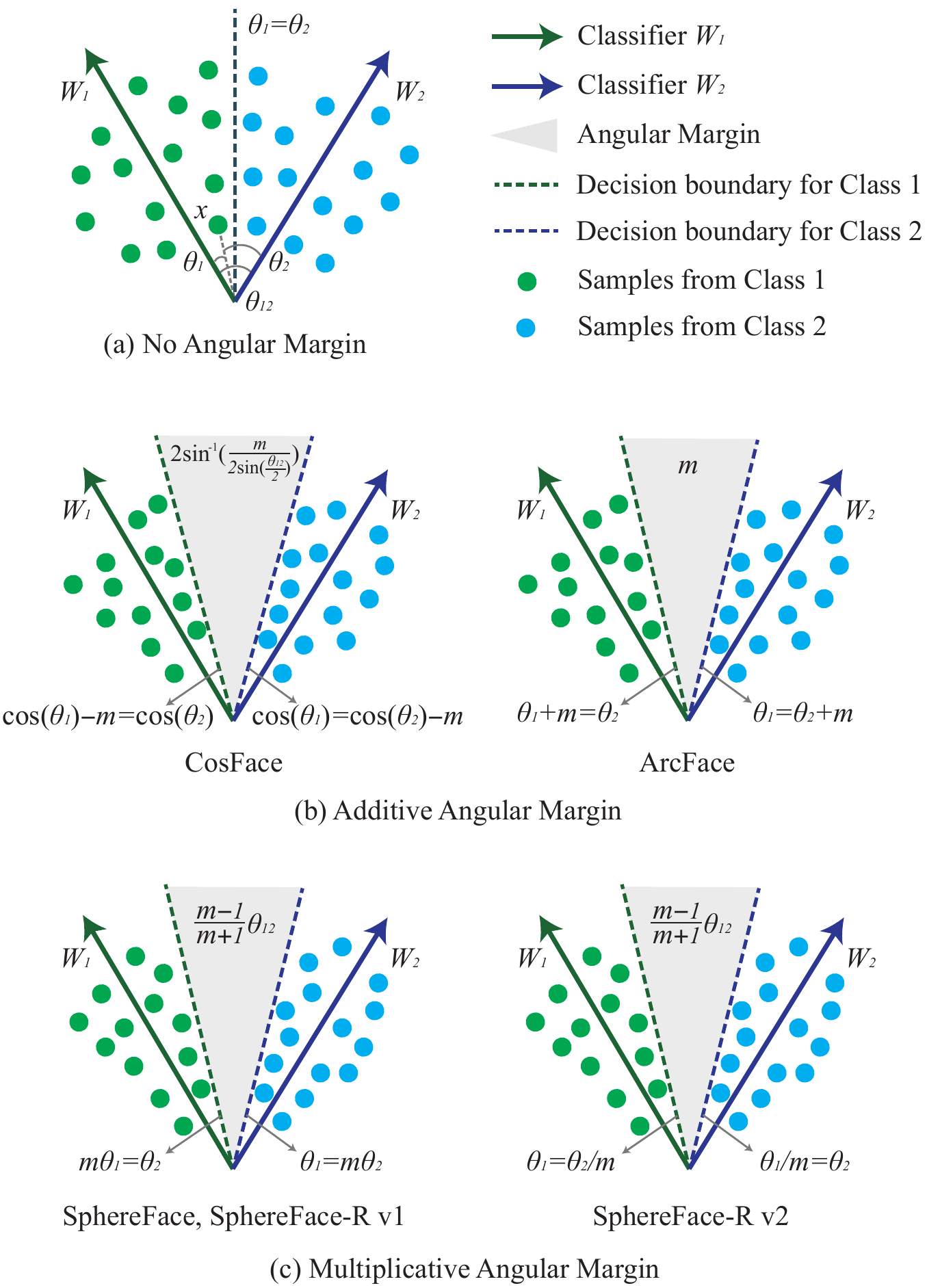}
 \vspace{-0.5mm}
  \caption{An intuitive comparison among no angular margin (\eg, \cite{wang2017normface,liu2017rethinking}), additive angular margin (CosFace~\cite{wang2018additive,wang2018cosface} and ArcFace~\cite{deng2019arcface}) and multiplicative angular margin (SphereFace, SphereFace-R v1 and SphereFace-R v2).}\label{compare-margin}
\end{figure}

As has been shown in Table~\ref{framework_case}, all the previous hyperspherical FR methods focus on designing a good target angular activation function. SphereFace adopts the same paradigm by constructing the following target angular function $\psi(\theta)$:
\begin{equation}\label{sphereface_psi}
    \psi(\theta)=(-1)^k\cos(m\theta)-2k,\ \ \ \theta\in\left[\frac{k\pi}{m}, \frac{(k+1)\pi}{m} \right]
\end{equation}
where $k\in[0,m-1]$ and $k$ is an integer. After ensuring $\eta(\theta)=\cos(\theta)$ and Eq.~\eqref{sphereface_psi}, Eq.~\eqref{new-softmax} becomes the objective function for the original SphereFace, as summarized in Table~\ref{framework_case}. The original SphereFace requires $m$ to be an integer, which is in fact unnecessary. $m$ can be any positive value larger than $1$. In order to improve training stability, our original SphereFace minimizes its objective function (with $m=4$) jointly with a standard softmax loss, which approximately yields an effective $m$ as $1.4$. Therefore, the target angular function in the original SphereFace can be simplified to
\begin{equation}\label{sphereface_psi_2}
\psi(\theta)=
\left\{
{\begin{array}{*{20}{l}}
\cos(m\theta) & \textnormal{if} & 0\leq\theta\leq\frac{\pi}{m}\\
-\cos(m\theta)-2 & \textnormal{if} & \frac{\pi}{m}<\theta\leq\pi
\end{array}} \right.
\end{equation}
where we usually use $m\in [1,2]$. One can easily verify that the characteristic function $\Delta(\theta)$ is always larger than zero with $\theta\in(0,\pi]$, so it satisfies the general principle to introduce large angular margin. Fig.~\ref{compare-margin} intuitively compares no angular margin, additive angular margin and multiplicative angular margin. The intuition of multiplicative margin can be understood from a simple binary classification example (with two classifiers $\bm{W}_1$ and $\bm{W}_2$). We consider Eq.~\eqref{generalized-softmax} with $\eta(\theta)=\cos(\theta)$ and $\psi(\theta)=\cos(\theta)$. For a sample $\bm{x}$, we need to require $\cos(\theta_1)>\cos(\theta_2)$ to correctly classify $\bm{x}$. But what if we instead require $ \cos(m\theta_1)>\cos(\theta_2)$ where $m>1$ in order to correctly classify $\bm{x}$? It is essentially making the decision more stringent than previous, because we require a lower bound\footnote{The inequality $\cos(\theta_1)>\cos(m\theta_1)$ holds if $\theta_1\in[0,\frac{\pi}{m}], m>1$.} of $\cos(\theta_1)$ to be larger than $\cos(\theta_2)$. The decision boundary for class 1 is $ \cos(m\theta_1)=\cos(\theta_2)$. Similarly, if we require $\cos(m\theta_2)>\cos(\theta_1)$ to correctly classify samples from class 2, the decision boundary for class 2 is $\cos(m\theta_2)=\cos(\theta_1)$. Suppose all training samples are correctly classified, such asymmetric decision boundaries will naturally produce an angular margin of size $\frac{m-1}{m+1}\theta_{12}$ where $\theta_{12}$ denotes the angle between $\bm{W}_1$ and $\bm{W}_2$. From angular perspective, correctly classifying $\bm{x}$ from identity 1 requires $\thickmuskip=2mu \theta_1<\frac{\theta_2}{m}$, while correctly classifying $\bm{x}$ from identity 2 requires $\thickmuskip=2mu \theta_2<\frac{\theta_1}{m}$. If $m>1$, both decision criteria are more difficult to achieve than the vanilla case without any angular margin (\ie, $\theta_1<\theta_2$ and $\theta_2<\theta_1$).

We can observe that Eq.~\eqref{sphereface_psi_2} can exactly match the intuition of multiplicative margin only when $\theta\in[0,\frac{\pi}{m}]$. When $\theta\in (\frac{\pi}{m},\pi]$, the same argument however will no longer hold. Although such a heuristic design can still empirically achieve large angular margin and work reasonably well, it may inevitably be less interpretable and also contribute to the training instability. The key to multiplicative angular margin is to guarantee that the equation $\psi(\theta)=\eta(m\theta)$ ($m>1$) always holds for $\theta\in[0,\pi]$. In order to better implement the intuition of multiplicative margin, we propose two different approaches, \ie, designing either a new target angular function $\psi(\theta)$ or a new non-target function $\eta(\theta)$. In Section~\ref{sfrv1}, we first follow the original idea of SphereFace to re-design a target angular function $\psi(\theta)$ which can better reflect the intuition of multiplicative margin. In Section~\ref{sfrv2}, we take a different approach by designing a new $\eta(\theta)$ which has a much simpler form yet can exactly match the intuition of multiplicative margin for $\theta\in[0,\pi]$. Section~\ref{section:cgd} proposes a useful implementation method to further stabilize training. Section~\ref{section:sfr-imp} gives implications and discussions.

\subsubsection{SphereFace-R v1: On Designing $\psi(\theta)$}\label{sfrv1}
Following the conventional way to design an angular margin loss~\cite{liu2017sphereface,wang2018additive,wang2018cosface,deng2019arcface}, we first focus on constructing a target angular function $\psi(\theta)$ based on the intuition of multiplicative margin. For $\theta\in[0,\frac{\pi}{m}]$, we can simply use $\psi(\theta)=\cos(m\theta)$ which is a monotonically decreasing function in $[0,\frac{\pi}{m}]$ and exactly implements the multiplicative angular margin. When $\theta>\frac{\pi}{m}$, SphereFace constructs a surrogate monotonically decreasing function to replace $\cos(m\theta)$, as specified in Eq.~\eqref{sphereface_psi_2}. However, this design of $\psi(\theta)$ in $[\frac{\pi}{m},\pi]$ does not follow the original intuition of multiplicative margin and may be sub-optimal. In order to better implement multiplicative margin in the entire domain of $[0,\pi]$, we propose the following target angular function:
\begin{equation}\label{sfr1_eq}
    \psi(\theta)=\cos\left( \min\Big\{ m,\frac{\pi}{\theta} \Big\} \cdot \theta \right)
\end{equation}
where $m$ is usually a prescribed positive constant. Eq.~\eqref{sfr1_eq} remains a monotonic function in $[0,\pi]$ and can be viewed as incorporating large angular margin with a dynamic multiplicative parameter $\min\{m,\frac{\pi}{\theta}\}$. For $\theta\in[0,\frac{\pi}{m}]$, Eq.~\eqref{sfr1_eq} is exactly the same as SphereFace and perfectly implements the multiplicative margin. For $\theta\in[\frac{\pi}{m},\pi]$, we consider a new multiplicative margin parameter $m'$ and the target angular function becomes $\psi(\theta)=\cos(m'\theta)$. In order to \emph{(i)} make $m'$ as large as possible and \emph{(ii)} make $\psi(\theta)$ a monotonic decreasing function where $m'\theta$ does not exceed $\pi$, we propose an adaptive decreasing strategy for $m'$: $m'=\frac{\pi}{\theta}$. Combining pieces, we end up with a dynamic multiplicative margin parameter $m'=\min\{m,\frac{\pi}{\theta}\}$. The non-target angular function is the same as SphereFace, \ie, $\eta(\theta)=\cos(\theta)$. Therefore, the multiplicative margin is implemented through $\psi(\theta)=\eta(m'\theta)$. The curve of the corresponding characteristic function $\Delta(\cdot)$ is given in Fig.~\ref{fig:delta_func}(b). More interestingly, we can observe that SphereFace-R v1 incorporates less angular margin to samples that are too easy or too hard (\ie, the target angle is around $0$ or $\pi$) and combines the largest angular margin to samples with medium hardness. We also compare SphereFace-R v1 with the other hyperspherical FR methods in Table~\ref{framework_case}.

Despite the well implemented multiplicative margin in Eq.~\eqref{sfr1_eq}, there is still a constraint on the effective multiplicative margin parameter $m'$, \ie, $m'\leq\frac{\pi}{\theta}$. Moreover, we have no consistent $m'$ in Eq.~\eqref{sfr1_eq} for samples with different target angle. It indicates that for an arbitrary sample whose target angle is within $[0,\pi]$, SphereFace-R v1 can not guarantee the same $m'$. To address this limitation, we propose to implement the multiplicative margin from the perspective of the non-target angular function rather than the target angular function, leading to SphereFace-R v2.

\subsubsection{SphereFace-R v2: On Designing $\eta(\theta)$}\label{sfrv2}

We consider how to design the non-target angular function $\eta(\theta)$ based on the intuition of multiplicative margin. To achieve $\psi(\theta)=\eta(m\theta)$ without changing the target angular function $\psi(\theta)$, we can naturally arrive at the following desired non-target angular function $\eta(\theta)$:
\begin{equation}\label{sfr2_eq}
    \eta(\theta)=\psi\left(\frac{\theta}{m}\right)=\cos\left(\frac{\theta}{m}\right)
\end{equation}
where $m$ is a prescribed positive constant and $\psi(\theta)=\cos(\theta)$. Compared to Eq.~\eqref{sfr1_eq} in SphereFace-R v1, Eq.~\eqref{sfr2_eq} is much simpler and more importantly, satisfies the property of $\psi(\theta)=\eta(m\theta)$ for $\theta\in[0,\pi]$ with a static $m$. While being extremely simple and conceptually appealing, SphereFace-R v2 can also exactly incorporate a static multiplicative angular margin. In contrast, SphereFace-R v1 is unable to induce a static multiplicative margin with $m>1$ and can only incorporate a dynamic multiplicative margin where the effective margin parameter has to be close to $1$ if $\theta$ is near $\pi$. More importantly, unlike SphereFace-R v1, there is no constraint for the size of the induced angular margin in SphereFace-R v2 and we can use any desirable $m\geq1$. 

From the corresponding characteristic function given in Fig.~\ref{fig:delta_func}(b), we can see that SphereFace-R v2 incorporates the smallest angular margin to easiest samples and the largest angular margin to samples with medium hardness. Unlike SphereFace-R v1 that introduces very small angular margin to hard samples, SphereFace-R v2 combines much larger angular margin to these samples. Therefore, SphereFace-R v1 and SphereFace-R v2 put different efforts on optimizing hard samples and may yield different generalizability.

To the best of our knowledge, SphereFace-R v2 is the very first method that introduces large angular margin through non-target angular functions. SphereFace-R v2 easily addresses the difficult problem of incorporating a static multiplicative margin by simply switching the design focus from target function to non-target function. We believe that this method provides an important and novel perspective on designing large angular margin losses.

\subsubsection{Characteristic Gradient Detachment}\label{section:cgd}

\begin{figure}[t]
  \centering
 \includegraphics[width=3.49in]{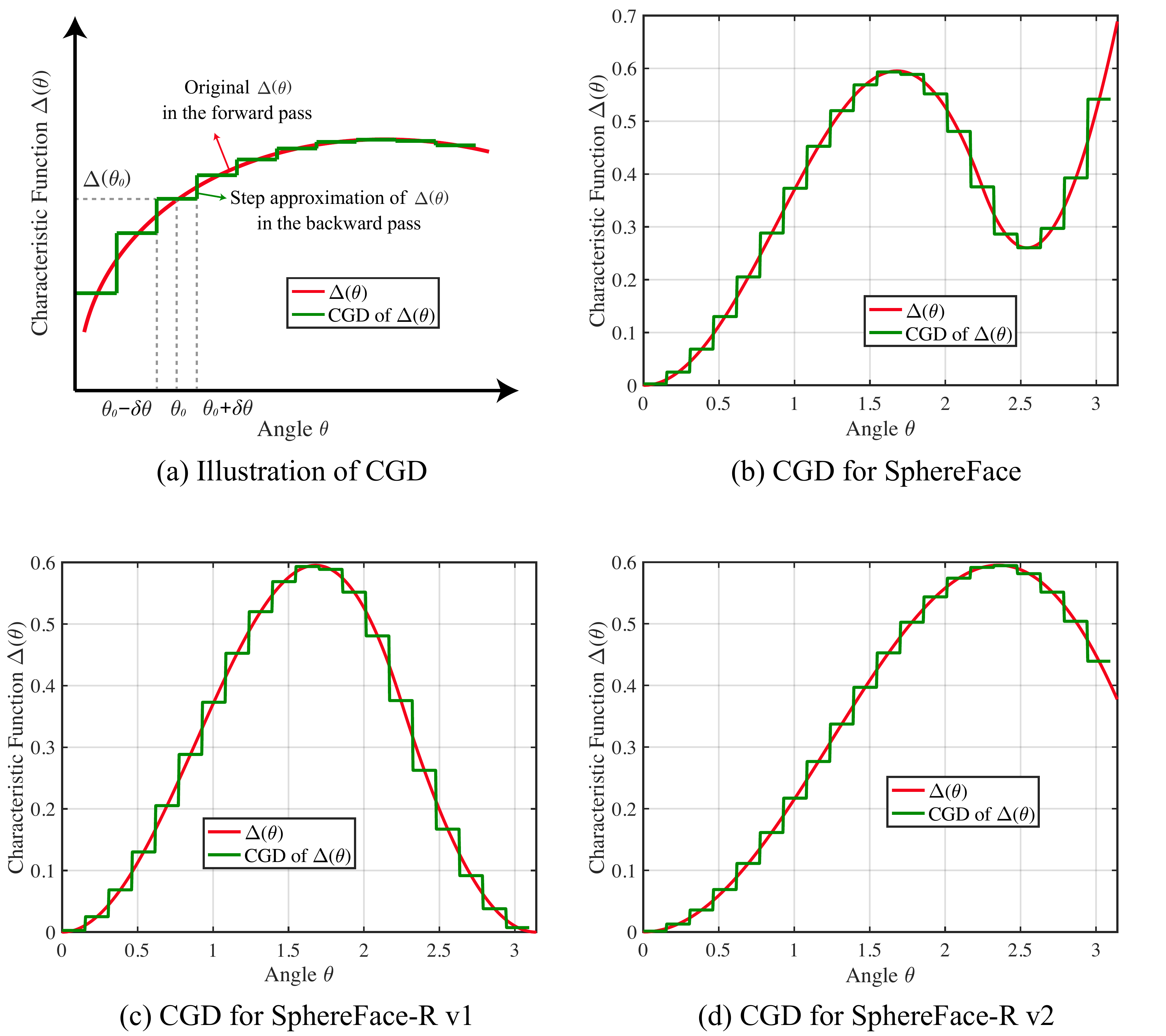}
  \vspace{-3mm}
  \caption{An illustration of the backward activation of CGD for SphereFace ($m=1.4$), SphereFace-R v1 ($m=1.4$) and SphereFace-R v2 ($m=1.4$). The green curves demonstrate the effect of CGD in the backward propagation, and the forward computation still follows the red curves (\ie, $\Delta(\theta)$) without any approximation.}\label{cgd_illu}
\end{figure}

In order to further stabilize training and improve performance, we introduce a simple and generic method -- characteristic gradient detachment for implementing our multiplicative margin. In general, the shape of the characteristic function $\Delta(\theta)$ determines the training stability and the convergence property. Empirically, we find that a characteristic function with simpler backward gradient computation typically leads to better training stability. For example, CosFace~\cite{wang2018additive,wang2018cosface} yields strong empirical training stability and its characteristic function is simply a positive constant with backward gradient as $0$. Inspired by such an observation, we aim to simplify the backward gradient computation for the characteristic function. To gain more intuitions, we first use Taylor expansion to decompose the characteristic function in the target angular function at an arbitrary angle $\theta_0\in(0,\pi)$ with a small angle deviation $\delta\theta$:
\begin{equation}\label{taylor_target}
\begin{aligned}
    \psi(\theta_0+\delta\theta)&=\eta(\theta_0+\delta\theta)-\Delta(\theta_0+\delta\theta)\\
    &=\eta(\theta_0+\delta\theta)-\big(\Delta(\theta_0)+\frac{\Delta'(\theta_0)}{1!}\delta\theta+\\
    &\ \ \ \ \ \ \ \cdots + \frac{\Delta^{(n)}(\theta_0)}{n!}(\delta\theta)^n+R_n(\delta\theta)\big)\\
\end{aligned}
\end{equation}
where $\Delta^{(n)}(\theta)$ denotes the $n$-th order derivative of $\Delta(\theta)$ and $R_n(\delta\theta)$ denotes the higher order infinitesimal of $(\delta\theta)^n$. When the characteristic function is more complex, then its Taylor expansion needs to have more terms to accurately represent it. This leads to more complex backward gradient computation. Motivated by the observation that simpler gradient computation often leads to better training stability, we propose to make an approximation to the characteristic function by removing some higher order terms in its Taylor expansion. Generally, we can remove any higher order Taylor expansion terms and it yields different backward gradients. Particularly, we draw inspirations from the constant characteristic function adopted in CosFace, and use the zero-order approximation for $\Delta(\theta_0+\delta\theta)$ in Eq.~\eqref{taylor_target}:
\begin{equation}\label{taylor_target2}
    \psi(\theta_0+\delta\theta)\approx \eta(\theta_0+\delta\theta)-\Delta(\theta_0)
\end{equation}
which is much simpler and robust to compute and gives the following approximate gradient for $\psi(\theta)$ at $\theta_0$:
\begin{equation}\label{taylor_gradient}
    \psi'(\theta_0)=\lim_{\delta\theta\rightarrow 0}\frac{\psi(\theta_0+\delta\theta)-\psi(\theta_0)}{\delta\theta}\approx \eta'(\theta_0)
\end{equation}
which naturally leads to the proposed CGD where we can simply apply gradient detachment to the characteristic function $\Delta(\theta)$. Specifically, we stop the gradient of the characteristic function with a detachment operator:
\begin{equation}\label{CGD}
    \psi(\theta)=\eta(\theta)-\textnormal{Detach}(\Delta(\theta))
\end{equation}
where $\textnormal{Detach}(\cdot)$ denotes the detachment operator that allows forward computation but stops the backward gradient propagation. This essentially means that we only need to compute the characteristic function in the forward pass and completely ignore it in the backward propagation. In order to avoid computing the gradient of the characteristic function, we substitute Eq.~\eqref{CGD} into the first line of Eq.~\eqref{new-softmax} and finally obtain the following loss function:
\begin{equation*}\label{softmax-v1}
    \mathcal{L}_{\textnormal{v1}} = \log\! \Big( 1 + \sum_{i\neq y}\!\exp\big(\norm{\bm{x}}(\eta(\theta_i)-\eta(\theta_y)+\textnormal{Detach}(\Delta(\theta_y)))\big) \Big)
\end{equation*}
which can be generally used for the cases where the target angular function is modified, such as SphereFace and SphereFace-R v1.
In the backward pass, CGD approximates the characteristic function of the multiplicative margin with a piece-wise function consisting of many constant functions, as illustrated in Fig.~\ref{cgd_illu}. Equivalently, CGD can also be viewed as a step function approximation to the characteristic function in the backward pass. We note that the approximation in CGD only exists in the backward direction and the forward computation is always identical to the CGD-free scenario. From a different perspective, CGD can be understood as interpreting the multiplicative margin with a dynamic CosFace-style additive margin (\ie, the effective margin parameter for the additive margin is dynamically dependent on the input target angle in CGD rather than being static in CosFace). The discussion above applies to both SphereFace and SphereFace-R v1, since they are modifying the target angular function. As a concrete example, applying CGD to SphereFace-R v1 yields the target angular function: $\psi(\theta)=\cos(\theta)-\textnormal{Detach}(\cos(\theta)-\cos(\min\{ m,\frac{\pi}{\theta} \}\cdot\theta))$ whose gradient is identical to CosFace.

For SphereFace-R v2 that modifies the non-target angular function, the derivation is similar except that we focus on approximating the gradient of the non-target function $\eta(\theta)$ instead of the target function $\psi(\theta)$. Therefore, we can similarly apply gradient detachment to the characteristic function in the non-target angular function:
\begin{equation}\label{CGD2}
    \eta(\theta)=\psi(\theta)+\textnormal{Detach}(\Delta(\theta)).
\end{equation}
After putting Eq.~\eqref{CGD2} into the first line of Eq.~\eqref{new-softmax}, we end up with the following general loss function for the cases that modify the non-target function:
\begin{equation*}\label{softmax-v2}
    \mathcal{L}_{\textnormal{v2}} = \log\! \Big( 1 + \sum_{i\neq y}\!\exp\big(\norm{\bm{x}}(\psi(\theta_i)-\psi(\theta_y)+\textnormal{Detach}(\Delta(\theta_i)))\big) \Big)
\end{equation*}
from which we can see that the key is to detach the gradients of the characteristic function. 
Therefore, applying CGD to SphereFace-R v2 yields the non-target function: $\eta(\theta)=\cos(\theta)+\textnormal{Detach}(\cos(\frac{\theta}{m})-\cos(\theta))$. The essence of CGD is to avoid computing the gradient of the characteristic function in the backward pass. As a simple generalization, we can consider higher-order Taylor approximation to the characteristic function instead of the zero-order approximation. Since CGD already yields satisfactory training stability and empirical performance, we will stick to it for simplicity.

In fact, CGD serves as a generally useful tool for implementing new types of angular margin and is not limited to SphereFace and SphereFace-R. For the backward propagation, CGD can approximate the characteristic function induced by any angular margin with a dynamic additive margin, and effectively stabilize the training.

\subsubsection{Implications and Discussions}\label{section:sfr-imp}

\textbf{Comparison between additive and multiplicative margin}. While Table~\ref{framework_case} provides a detailed comparison between additive and multiplicative margin, the fundamental difference between them is on a conceptual level. Additive margin is introduced by adding or subtracting a parameter to the target function so that the characteristic function $\Delta(\theta)$ can be larger than zero in most cases. Specifically, this parameter can be either inside~\cite{deng2019arcface} or outside~\cite{wang2018additive,wang2018cosface} the cosine function. In contrast, multiplicative margin is achieved by multiplying a parameter to the target or non-target function so that $\Delta(\theta)$ is larger than zero. It is also possible that a multiplicative margin loss and an additive margin loss lead to the same characteristic function, and they may be technically the same loss. Therefore, their difference is determined by the specific intuition that guides the loss design.

\vspace{0.7mm}
\noindent\textbf{Generality of multiplicative margin}. SphereFace-R v1 and v2 demonstrate two different strategies to incorporate multiplicative angular margin, showing the existence of many feasible designs to achieve multiplicative margin. In fact, the exact form of the loss function is not crucial and the core of multiplicative margin lies in the spirit of multiplying a factor to ensure the characteristic function to be larger than zero. Following such a spirit, there are likely many potential loss designs that can work as well as ours.

\vspace{0.7mm}
\noindent\textbf{Comparison between SphereFace and SphereFace-R}. It is easy to see that SphereFace employs a surrogate characteristic function to implement the multiplicative margin and does not follow the intuition of multiplicative margin for $\theta\in[\frac{\pi}{m},\pi]$ where $m>1$. In contrast, both SphereFace-R v1 and v2 exactly follow the intuition of multiplicative margin in the entire domain of $[0,\pi]$. SphereFace-R v1 implements a dynamic multiplicative margin (\ie, the effective margin parameter varies depending on the training sample), while SphereFace-R v2 implements a static one (\ie, the effective margin parameter stays the same for all training samples). Moreover, SphereFace and SphereFace-R use different effective margin parameters for samples with different hardness. Both SphereFace and SphereFace-R strictly satisfy the general principle in Eq.~\eqref{margin-principle} to achieve large angular margin, validating the effectiveness of our proposed principle.

\vspace{0.7mm}
\noindent\textbf{Jointly designing target and non-target functions}. Because SphereFace-R v1 focuses on the target angular function and SphereFace-R v2 focuses on the non-target angular function, it is natural to consider to simultaneously design the target and non-target angular functions. For example, Eq.~\eqref{sfr1_eq} and Eq.~\eqref{sfr2_eq} can be easily used together and the resulting characteristic function is simply the combination of both. More interestingly, it is not necessary for both target and non-target functions to use the cosine-based design. We can simply use a linear function as the target and non-target functions, as proposed in \cite{liu2017deep}. It can effectively alleviate some design constraints caused by the periodicity of cosine function. Jointly designing the target and non-target functions can greatly enlarge the search space of the characteristic function and may lead to a better multiplicative margin loss.

\vspace{0.7mm}
\noindent\textbf{Beyond additive and multiplicative margin}. There are many more alternative types of angular margin other than the additive and multiplicative ones. For example, we can also use the exponential function to achieve $\Delta(\theta)>0$. Specifically, we use $\eta(\theta)=\cos(2\theta)$ and $\psi(\theta)=\cos^m(2\theta)$, where $m>1$ is the margin parameter and larger $m$ gives larger angular margin. Alternatively, we can also combine additive and multiplicative margin as $\eta(\theta)=\cos(\theta),\psi(\theta)=\cos(m_1\theta+m_2)-m_3$. It remains an open problem to design a simple yet well-performing angular margin.

\subsection{Feature Magnitude}\label{section:mag}

SphereFace~\cite{liu2017sphereface} originally does not use feature normalization, because the feature magnitude does not affect the angular decision boundary. \cite{wang2017normface,wang2018additive,wang2018cosface,deng2019arcface} show that feature normalization can ease the difficulty of minimizing angular margin losses and greatly improve the training stability. Despite being effective to stabilize training, feature normalization inevitably loses useful information about individual samples (\eg, image quality). Existing hyperspherical FR methods either preserve the feature magnitude in the loss function~\cite{liu2016large,liu2017sphereface,liu2018learning} or normalize the feature magnitude to constant $s$~\cite{wang2017normface,liu2017rethinking}. To explore whether feature magnitude can be beneficial to generalization, we systematically study SphereFace and SphereFace-R under NFN and HFN. Moreover, we consider a soft feature normalization method which effectively unifies NFN and HFN and serves as an interpolation between both.

\vspace{0.7mm}
\noindent\textbf{Hard feature normalization}. HFN becomes a default component in current hyperspherical FR methods~\cite{wang2018additive,wang2018cosface,deng2019arcface,sun2020circle,huang2020curricularface}. By normalizing the feature $\bm{x}$ to a constant $s$, the objective function value will merely depend on the angles between $\bm{x}$ and the classifiers $\bm{W}_i,\forall i$. In order to perform such a hard normalization on the feature $\bm{x}$, we parameterize the original $\bm{x}$ in Eq.~\eqref{new-softmax} with $s\frac{\bm{x}}{\|\bm{x}\|}$ and arrive at Eq.~\eqref{normalized-softmax}. Since $s$ is a prescribed constant, it is equivalent to normalizing all the features to a hypersphere with radius $s$.

\vspace{0.7mm}
\noindent\textbf{Soft feature normalization as an interpolation}. We consider the soft feature normalization that interpolates between FN-free learning and HFN. Specifically, besides the original loss, we combine an additional regularization term to constrain the feature magnitude:
\begin{equation}
    \mathcal{L}_{\textnormal{SFN}}=t\cdot\big{\|}\|{\bm{x}}\|-s\big{\|}^2
\end{equation}
where $t$ is a hyperparameter that controls the regularization strength and $s$ is a prescribed feature magnitude that serves a similar role to HFN. When $t=0$, SFN reduces to FN-free learning. When $t=+\infty$, SFN reduces to HFN. Therefore, SFN can be viewed as an interpolation between FN-free learning and HFN. 
SFN has also been studied in \cite{zheng2018ring}.

SFN can make use of the instance-level information encoded in feature magnitude during training while still encouraging a feature normalization effects. Moreover, the difference between SFN and HFN can be viewed as using different optimization techniques to constrain the feature norm to a prescribed constant. HFN has the flavor of projected gradient descent where the solution will be projected to the feasible region to satisfy some constraint. In contrast, SFN is essentially a Lagrangian relaxation of the original problem where the feature norm is constrained. Therefore, their empirical performance could be quite different in practice, even if they share the same optimization target.

\vspace{0.7mm}
\noindent\textbf{Dynamic feature magnitude}. In contrast to HFN that uses a static feature magnitude, both FN-free learning and SFN can be viewed as a dynamic (data-dependent) way to control the feature magnitude. Moreover, there exist many other strategies that can dynamically control the feature magnitude to improve the empirical performance, such as \cite{zhang2019adacos}.

\subsection{Discussions and Open Problems}\label{section:diss}

\textbf{Optimal design of characteristic function}. It is clear that the characteristic function is the key to large angular margin, but is there an optimal characteristic function? The answer to this question remains open. We argue that the optimal design of characteristic function should be dynamic and depends on the specific dataset, the network architecture, the optimizer, the stage of training (\ie, the weights of the network), etc. Current studies on hyperspherical FR still focuses on a static characteristic function. \cite{wang2020loss,li2019lfs} explore an automatic way to learn a characteristic function from data, but those learned characteristic functions are still static ones and do not lead to a significant performance gain. \cite{meng2021magface} combines the sample quality to hyperspherical FR through a customized characteristic function that is dependent on the feature magnitude. How to design or learn a better characteristic function that is dynamically dependent on the data and also easy to optimize remains a huge challenge. Moreover, the underlying mechanism that determines the performance of a characteristic function stays a mystery and needs to be understood both empirically and theoretically.

\vspace{0.7mm}
\noindent\textbf{Making better use of feature magnitude}. In this paper, we have not considered to incorporate feature magnitude to testing and still stick to the cosine similarity for comparing pairs. However, it remains an interesting open problem whether it will be more beneficial to combine feature magnitude back to the similarity score (especially FN-free learning or SFN is used). We consider a generalized form of similarity score as $\mathcal{S}(\bm{x}_1,\bm{x}_2)=g(\norm{\bm{x}_1},\norm{\bm{x}_2})\cdot\cos(\theta_{1,2})$ where $\bm{x}_1,\bm{x}_2$ are deep features of two input samples, $\theta_{1,2}$ is the angle between $\bm{x}_1$ and $\bm{x}_2$, and $g(\|{\bm{x}_1}\|,\|{\bm{x}_2}\|)$ denotes a function with the norm of $\bm{x}_1$ and $\bm{x}_2$ as input. We may require the function $g(\|{\bm{x}_1}\|,\|{\bm{x}_2}\|)$ to have a few properties: (i) permutation invariance: $g(\|{\bm{x}_1}\|,\|{\bm{x}_2}\|)=g(\|{\bm{x}_2}\|,\|{\bm{x}_1}\|)$ and (ii) adjustable magnitude augmentation. As a concrete example, we could use $g(\norm{\bm{x}_1},\norm{\bm{x}_2})=\norm{\bm{x}_1}^t\cdot\norm{\bm{x}_2}^t$ where $t$ adjusts the augmentation strength of feature magnitude. $g(\norm{\bm{x}_1},\norm{\bm{x}_2})=1$ reduces to the cosine similarity score.  In general, how to design a good $g$ is not clear and remains to be explored in future endeavours. 

\section{A Unified Characterization of Loss Functions in Hyperspherical Face Recognition}

\begin{figure}[t]
  \centering
 \includegraphics[width=3.35in]{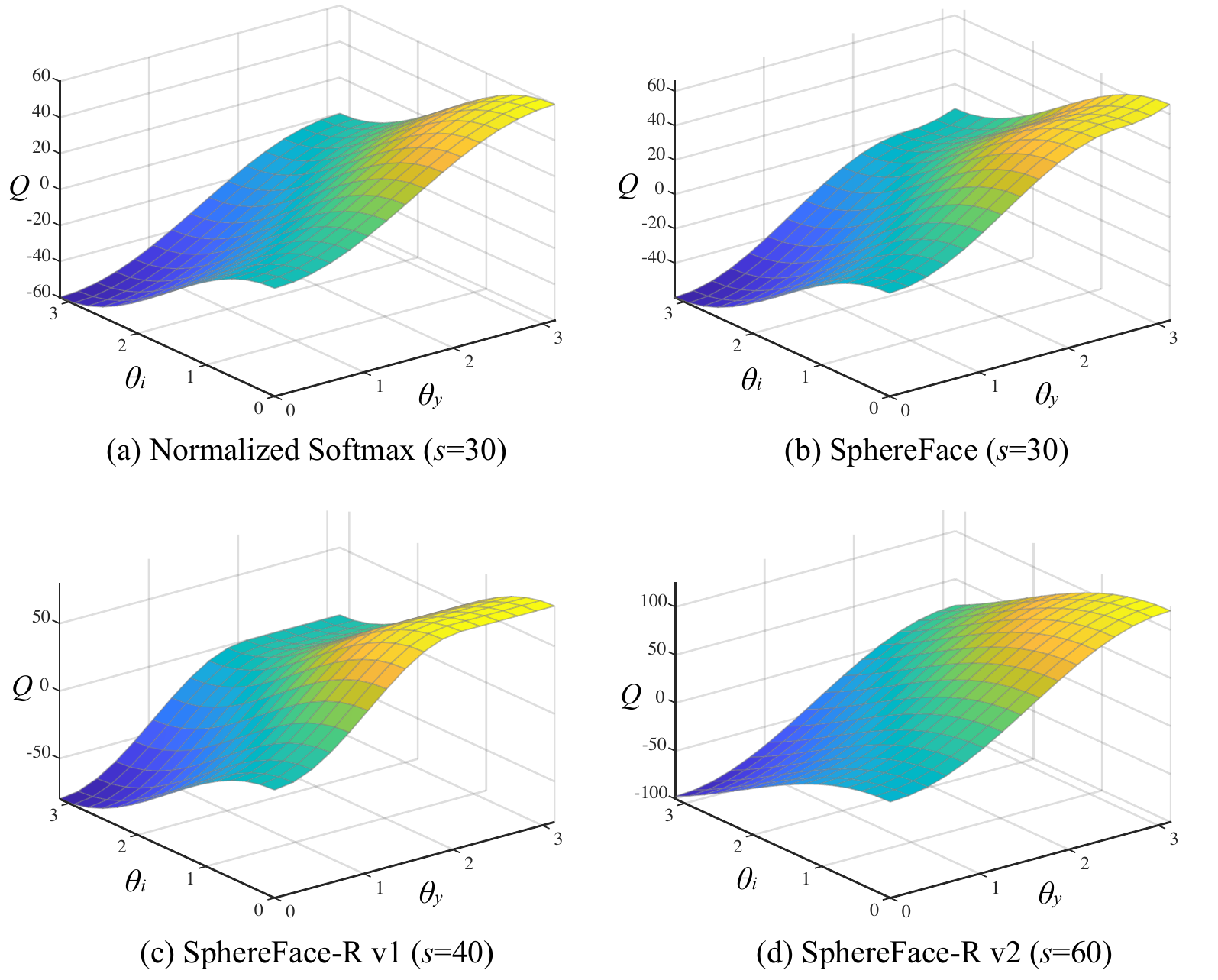}
 \vspace{-0.5mm}
  \caption{A comparison of loss characteristics among normalized softmax, SphereFace, SphereFace-R v1 and SphereFace-R v2.}\label{q_func}
\end{figure}

In this section, we take a closer look at what characterizes hyperspherical face recognition. As our unified framework in Section~\ref{unified_framework} discusses, a feature normalization strategy, a (non-)target angular function and a characteristic function 
can fully determine the loss function of a hyperspherical FR method. Particularly, the characteristic function $\Delta(\cdot)$ controls the property of the induced angular margin (\eg, size, training stability). It does not consider the feature magnitude and only focuses on the difference between target and non-target function. Here we take a step further by showing a unified way to characterize the loss function as a whole. Specifically, we have the following general form of the loss function for hyperspherical FR:
\begin{equation}\label{labeled-normalized-softmax}
\begin{aligned}
    \mathcal{L}_{s} &= \log \Big( 1 + \sum_{i\neq y}\exp\big(\underbrace{s\cdot(\eta(\theta_i)-\eta(\theta_y)+\Delta(\theta_y))}_{:=Q(\theta_y,\theta_i,s,m)}\big) \Big)\\
    &=\log \Big( 1 + \sum_{i\neq y}\exp\big( Q(\theta_y,\theta_i,s,m)\big)\Big)
\end{aligned}
\end{equation}
where we define $Q(\theta_y,\theta_i,s,m)$ as the loss characteristics that fully determine how the loss function behaves. We compare the loss characteristics among normalized softmax~\cite{wang2017normface}, SphereFace and two variants of SphereFace-R in Fig.~\ref{q_func}. Although we show that the loss characteristics can fully determine the loss function, the underlying mechanisms of how the loss characteristics can affect the performance are largely unclear and remain to be investigated. Typically, $s$ and $m$ jointly specify the loss characteristics, and their roles could be partially coupled, which is also empirically observed in our ablation study.

From a back-propagation perspective, we have the gradient of $\mathcal{L}_s$ (\emph{w.r.t.} either $\bm{x}$ or $\bm{W}_i,\forall i$) as
\begin{equation}\label{gradient-normalized-softmax}
    \!\!\mathcal{L}_{s}' \!=\! \sum_{i\neq y}\frac{\exp(Q(\theta_y,\theta_i,s,m))}{1\!+\!\sum_{i\neq y} \exp(Q(\theta_y,\theta_i,s,m))} \cdot Q'(\theta_y,\theta_i,s,m)
\end{equation}
where $Q'(\theta_y,\theta_i,s,m)$ denotes the gradient of loss characteristics. Quite interestingly, if we apply CGD to Eq.~\eqref{labeled-normalized-softmax}, then $Q'(\theta_y,\theta_i,s,m)$ for all hyperspherical FR methods will immediately become identical to that of the normalized softmax loss~\cite{wang2017normface}. The only critical difference lies in the weighting factor $\rho_i=\frac{\exp(Q(\theta_y,\theta_i,s,m))}{1+\sum_{i\neq y} \exp(Q(\theta_y,\theta_i,s,m))}$ in Eq.~\eqref{gradient-normalized-softmax}, where $Q(\theta_y,\theta_i,s,m)$ varies for different loss functions in the forward pass. This finding suggests that once CGD is applied, the gradient of every loss function in hyperspherical FR can be viewed as a particular weighting strategy to combine $Q'(\theta_y,\theta_i,s,m)$ of different $i\neq y$. In other words, only the weighting factors $\rho_i,\forall i$ in the gradient $\mathcal{L}_s'=\sum_{i\neq y}\rho_i\cdot Q'(\theta_y,\theta_i,s,m)$ will differ for different loss functions. Therefore, the design space for loss functions can be switched from finding $Q(\theta_y,\theta_i,s,m)$ to finding a weighting strategy for combining the gradients $Q'(\theta_y,\theta_i,s,m),\forall i\neq y$. Such a gradient weighting perspective reveals that searching for suitable $m$ and $m$ is equivalent to designing a good gradient weighting strategy. This may open a brand new gate to gain deeper understandings towards hyperspherical FR. Moreover, we believe that our novel loss characterization reformulation in Eq.~\eqref{labeled-normalized-softmax} and Eq.~\eqref{gradient-normalized-softmax} may inspire more effective designs for loss functions in hyperspherical FR.

\section{Experiments and Results}

In this section, we present comprehensive experiments to explore the properties of the SphereFace family. Experimental setup is introduced in Section~\ref{exp_detail}. We perform ablation studies in Section~\ref{sec:exp_vgg2} and Section~\ref{sec:exp_ms1m} to investigate different variants of SphereFace and their hyperparameters. In Section~\ref{exp_benchmark}, we evaluate our methods on the large-scale benchmarks and compare to ohter state-of-art methods.

\subsection{Implementation Details}\label{exp_detail}

\noindent\textbf{Preprocessing}. Each face image is cropped based on the five face landmarks (\ie, left eye, right eye, nasal tip, left mouth corner, and right mouth corner) detected by MTCNN~\cite{zhang2016joint} {\color{modify}and RetinaFace~\cite{deng2020retinaface}} using similarity transformation. The size of the cropped image is set to $112 \times 112$, and each RGB pixel ($[0,255]$) is normalized to $[-1,1]$.

\vspace{0.5mm}

\noindent\textbf{CNNs}. The SphereFace Networks (SFNets) that are initially proposed in \cite{liu2017sphereface} are used as the backbone in our experiments. Slightly different from \cite{liu2017sphereface}, we equip SFNets with batch normalization (BN) \cite{ioffe2015batch} to facilitate the model optimization. {\color{modify}For better comparison to existing methods, we also evaluate our models with IResNet-100~\cite{deng2019arcface} which is a 100-layer modified ResNet. The affine parameters in the last BN layer are enabled when NFN and SFN are used.} We use SFNet-20 and SFNet-64 in the ablation and exploration, while SFNet-64 {\color{modify}and IResNet-100} are adopted in large-scale benchmarks to achieve state-of-the-art performance.

\vspace{0.5mm}

\begin{table}[t]
	\centering
	\setlength{\tabcolsep}{3.8pt}
	\renewcommand{\arraystretch}{1.3}
	\caption{Statistics for the used datasets.}
	\vspace{-2mm}
	\begin{tabular}{l|c|c|c}
		Dataset & \# of ID & \# of images & Split \\
	    \shline
	    VGGFace2 \cite{cao2018vggface2} & 8.6K & 3.1M & train \\
	    MS-Celeb-1M \cite{guo2016ms} & 86K & 5.8M & train \\
	     \hline
	    LFW \cite{huang2007labeled} & 5,749 & 13,233 & validation \\
	    AgeDB-30 \cite{moschoglou2017agedb}& 568 & 16,488 & validation \\
	    CALFW \cite{zheng2018cross} & 5,749 & 11,652 & validation \\
	    CPLFW \cite{zheng2017cross} & 5,749 & 12,174 & validation \\
	    CFP \cite{sengupta2016frontal} & 500 & 7,000 & validation \\
	    VGGFace2\_test \cite{cao2018vggface2} & 500 & 173k & validation \\
		\hline
		IJB-B \cite{whitelam2017iarpa} & 1,845 & 76.8K & test \\
		IJB-C \cite{maze2018iarpa} & 3,531 & 148.8K & test \\
		MegaFace (probe) \cite{Kemelmacher-Shlizerman_2016_CVPR} & 530 & 3,530 & test \\
		MegaFace (distractor) \cite{miller2015megaface} & 690K & 1M & test \\
		\end{tabular}
	\label{datasets}
\end{table}

\begin{table}[t]
    \centering
	\setlength{\tabcolsep}{3.5pt}
	\renewcommand{\arraystretch}{1.25}
	\caption{Varying $m$ for no feature normalization on VGGFace2 ($\%$).}
	\vspace{-2mm}
	\begin{tabular}{c|c|c|c}
		$m$ & SphereFace & SphereFace-R v1 & SphereFace-R v2 \\
		\shline
		1.1 & 53.09 & 55.18 & 52.71 \\
        1.2 & \textbf{55.36} & \textbf{55.97} & \textbf{56.08} \\
        1.3 & 55.32 & 51.11 & 50.19 \\
        1.4 & 44.95 & 43.04 & 37.78 \\
        1.5 & 34.23 & 31.54 & 30.94 
		\end{tabular} \label{tab:nfn}
\vspace{-2mm}
\end{table}

\begin{table*}[t]
    \centering
	\setlength{\tabcolsep}{3.5pt}
	\renewcommand{\arraystretch}{1.25}
	\caption{Grid searching for $m$ and $s$ with hard feature normalization on VGGFace2. Results are in $\%$ and higher number indicates better performance.}
	\vspace{-4mm}
	\subfloat[Varying $m$ and $s$ for SphereFace. \label{tab:hfn-v0}] {	\begin{tabular}{c|c|c|c|c|c}
	\backslashbox{$m$}{$s$} & 20 & 30 & 40 & 50 & 60 \\
	\shline
	1.1 & 51.97 & 53.28 & 55.50 & 53.78 & 52.06 \\
	1.2 & \textbf{53.16} & \underline{\textbf{61.37}} & 57.22 & 54.85 & 56.12 \\
	1.3 & 48.28 & 60.08 & \textbf{60.89} & \textbf{60.37} & \textbf{57.09} \\
	1.4 & 42.32 & 58.63 & 59.30 & 57.39 & 53.94 \\
	1.5 & 37.20 & 58.04 & 59.35 & 54.74 & 53.43 \\
	1.6 & 38.62 & 47.81 & 53.53 & 49.77 & 53.30 
	\end{tabular}} \hspace{3mm}
	\subfloat[Varying $m$ and $s$ for SphereFace-R v1. \label{tab:hfn-v1}]{\begin{tabular}{c|c|c|c|c|c}
	\backslashbox{$m$}{$s$} & 20 & 30 & 40 & 50 & 60 \\
	\shline
	1.1 & \textbf{50.88} & 57.10 & 54.12 & 51.84 & 53.37 \\
	1.2 & 49.68 & 57.21 & 56.30 & 51.99 & 52.17 \\
	1.3 & 45.35 & \textbf{59.15} & 56.23 & 56.92 & 55.02 \\
	1.4 & 45.83 & 53.38 & 58.05 & 55.79 & 56.15 \\
	1.5 & 35.91 & 55.00 & \underline{\textbf{60.45}} & \textbf{58.29} & \textbf{58.95} \\
	1.6 & 33.40 & 47.70 & 53.75 & 52.28 & 50.92
	\end{tabular}} \hspace{4mm}
	\subfloat[Varying $m$ and $s$ for SphereFace-R v2. \label{tab:hfn-v2}]{\begin{tabular}{c|c|c|c|c|c}
	\backslashbox{$m$}{$s$} & 30 & 40 & 50 & 60 & 70 \\
	\shline
	1.1 & 54.62 & 54.69 & 50.52 & 52.56 & 51.33 \\
	1.2 & \textbf{55.59} & \textbf{60.75} & 58.30 & 54.55 & 56.88 \\
	1.3 & 44.12 & 55.67 & \textbf{61.07} & 57.34 & 57.15 \\
	1.4 & 35.01 & 48.42 & 59.31 & \underline{\textbf{62.72}} & \textbf{58.12} \\
	1.5 & 32.71 & 42.50 & 46.98 & 55.54 & 56.81 \\
	1.6 & 25.90 & 30.83 & 42.88 & 49.83 & 51.40
	\end{tabular}}
\vspace{-3mm}\label{tab:hfn_ablation}
\end{table*}

\begin{table}[t]
    \centering
	\setlength{\tabcolsep}{3.5pt}
	\renewcommand{\arraystretch}{1.2}
	\caption{Varying $t$ for soft feature normalization on VGGFace2 ($\%$).}
	\vspace{-2mm}
	\begin{tabular}{c|c|c|c}
		\multirow{2}{*}{$t$} & SphereFace &        SphereFace-R v1 & SphereFace-R v2 \\
	     & {\scriptsize($m=1.2$, $s=30$)} & {\scriptsize($m=1.5$, $s=40$)} & {\scriptsize($m=1.4$, $s=60$)} \\
		\shline
		0.05 & 57.08 & 39.43 & 55.53 \\
        0.1 & \textbf{62.40} & 43.96 & 60.61 \\
        0.2 & 61.24 & 54.40 & 60.30 \\
        0.5 & 58.68 & \textbf{61.37} & \textbf{61.21} \\
        1.0 & 57.10 & 60.59 & 58.34
		\end{tabular} \label{tab:sfn}
\vspace{-2mm}
\end{table}

\begin{table}[t]
    \centering
	\setlength{\tabcolsep}{3.1pt}
	\renewcommand{\arraystretch}{1.2}
	\caption{Ablation of CGD for different FN strategies on VGGFace2 ($\%$).}
	\vspace{-2mm}
	\begin{tabular}{c|c|c|c|c}
		\multirow{2}{*}{FN} & \multirow{2}{*}{CGD} & SphereFace &        SphereFace-R v1 & SphereFace-R v2 \\
	     & & {\scriptsize($m=1.2$, $s=30$)} &  {\scriptsize($m=1.5$, $s=40$)} &  {\scriptsize($m=1.4$, $s=60$)} \\
		\shline
		\multirow{2}{*}{NFN} & \xmark  & 47.91 & 49.92 & 49.41 \\
		 & \cmark & \textbf{55.36} & \textbf{55.97} & \textbf{56.08}  \\
		\hline
		\multirow{2}{*}{HFN} & \xmark  & 46.02 & 28.78 & 36.61\\
		 & \cmark & \textbf{61.37} & \textbf{60.45} & \textbf{62.72}  \\
		\hline
		\multirow{2}{*}{SFN} & \xmark  & 47.81 & 25.26 & 40.52 \\
		 & \cmark & \textbf{62.40} & \textbf{61.37} & \textbf{61.21}  \\
		\end{tabular} \label{tab:CGD}
\vspace{-2mm}
\end{table}

\begin{table}[t]
    \centering
	\setlength{\tabcolsep}{3.5pt}
	\renewcommand{\arraystretch}{1.2}
	\caption{Varying $m$ with no feature normalization on MS-Celeb-1M ($\%$).}
	\vspace{-2mm}
	\begin{tabular}{c|c|c|c}
		$m$ & SphereFace &  SphereFace-R v1 & SphereFace-R v2 \\
		\shline
         1.0 & 76.50 & 75.12 & 78.98 \\
         1.1 & \textbf{79.89} & 82.52 & 79.94 \\
         1.2 & 49.64 & \textbf{84.66} & \textbf{84.78} \\
         1.3 & - & 0.33 & 69.95 \\
		\end{tabular} \label{tab:ms1m-nfn}
\vspace{-2mm}
\end{table}

\begin{table}[t]
    \centering
	\setlength{\tabcolsep}{3.5pt}
	\renewcommand{\arraystretch}{1.2}
	\caption{Varying $m$ with hard feature normalization on MS-Celeb-1M ($\%$).}
	\vspace{-2mm}
	\begin{tabular}{c|c|c|c|c|c|c}
		\multirow{2}{*}{$m$} & \multicolumn{2}{c|}{SphereFace} &  \multicolumn{2}{c|}{SphereFace-R v1} & \multicolumn{2}{c}{SphereFace-R v2} \\
	     & $s=32$ & $s=64$ & $s=32$ & $s=64$ & $s=64$ & $s=96$ \\
		\shline
          1.4 & 87.32 & - & 89.42 & - & 86.88 & - \\
         1.5 & 88.82 & - & 90.19 & - & \underline{\textbf{91.58}} & - \\
         1.6 & 88.97 & 89.15 & \underline{\textbf{90.78}} & 86.24 & 88.74 & 88.48 \\
         1.7 & \underline{\textbf{91.03}} & 89.30 & 89.79 & 88.89 & 88.80 & 89.36 \\
         1.8 & 89.83 & \textbf{90.53} & 88.53 & 88.89 & 88.54 & 88.98 \\
         1.9 & - & 89.58 & - & \textbf{89.61} & - & \textbf{89.71} \\
         2.0 & - & 88.50 & - & 87.52 & - & 89.01 \\
		\end{tabular} \label{tab:ms1m-hfn}
\vspace{-2mm}
\end{table}

\begin{table}[!t]
    \centering
	\setlength{\tabcolsep}{3.0pt}
	\renewcommand{\arraystretch}{1.2}
	\caption{Varying $t$ with soft feature normalization on MS-Celeb-1M ($\%$).}
	\vspace{-2mm}
	{
	\begin{tabular}{c|c|c}
	\multirow{2}{*}{$t$} & SphereFace & SphereFace-R v1 \\
	  & {\scriptsize($m=1.7$, $s=32$)} & {\scriptsize($m=1.6$, $s=32$)} \\
	\shline
	0.6 & 54.30 & 81.27 \\
	0.8 & 85.40 & 81.24 \\
	1.0 & \textbf{87.70} & \textbf{87.19} \\
	1.2 & 85.00 & 87.15 \\
	1.4 & 78.52 & 83.67  
	\end{tabular}}  \hspace{2mm} {\begin{tabular}{c|c}
	\multirow{2}{*}{$t$} & SphereFace-R v2 \\
	   & {\scriptsize($m=1.5$, $s=64$)} \\
	\shline
	0.1 & 78.18 \\
	0.2 & \textbf{86.10} \\
	0.4 & 83.34 \\
	0.6 & 85.50 \\
	0.8 & 81.75
	\end{tabular}}
\vspace{-2mm}\label{tab:sfn_ms1m}
\end{table}

\noindent\textbf{Training}. The training images are horizontally flipped for data augmentation. We train all the models on two popular training dataset: VGGFace2~\cite{cao2018vggface2} (3.1M images from 8.6K IDs) and MS-Celeb-1M (5.8M images from 86K IDs{\color{modify}, also called MS1M-V2, the cleaned version of MS-Celeb-1M used in \cite{deng2019arcface}}). Detailed statistics of these datasets are given in Table~\ref{datasets}. In our experiments, all the models are optimized using stochastic gradient descent with momentum 0.9. For VGGFace2, we train on 2 GPUs for 80k iterations, with a learning rate of 0.1 which is decreased by 10 at the 40k and 60k iteration. For MS-Celeb-1M, we train on 4 GPUs for 240k iterations. The learning rate is initialized as 0.1 and decreased by 10 at the iteration of 100k, 180k, and 220k.
\vspace{0.5mm}

\noindent\textbf{Testing}. We strictly follow the specific protocol provided in each dataset for evaluation. Table~\ref{datasets} shows the statistics of the testing sets. Given a face image, we extract two 512-dimensional embeddings from the original image and its horizontally flipped version, respectively. The final embedding is obtained by averaging the two. The scoring method is cosine similarity. The nearest neighbor classifier and thresholding are used for face identification and verification, respectively. To reduce the randomness, 5 models from the last 10k iterations will be used in testing and their averaged results are reported. Specifically, we evaluate the models at the iteration of 72k, 74k, 76k, 78k and 80k for VGGFace2, and 232k, 234k, 236k, 238k and 240k for MS-Celeb-1M. 

\subsection{Ablation and Exploration on VGGFace2} \label{sec:exp_vgg2}
The validation set is a combination of multiple datasets, including LFW, AgeDB-30, CALFW, CPLFW, CFP-FP, CFP-FF, and VGG2-FP. The statistics of these datasets are summarized in Table~\ref{datasets}. In total, there are 43,000 testing pairs (21,500 positive pairs, and 21,500 negative pairs). The performance of a model is measured by the area under the ROC curve (AUC). Since it is important for a face recognition system to avoid false positives, we use AUC-$x$ \cite{barra2018footprints} as the metric, which integrates up to false positive rate of $x$ ($x\in [0, 1]$). We find that $x=0.0005$ achieves the best trade-off between stability and effectiveness. Since VGGFace2 is a relatively small training set (8.6K subjects \cite{cao2018vggface2}), we use the SFNet-20 model as the backbone in this section.

\subsubsection{No Feature Normalization}
We start by exploring SphereFace and SphereFace-R without feature normalization. Since there is only one effective hyperparameter (\ie, the margin $m$), it is easy to find the optimal setting. We compare SphereFace, SphereFace-R v1 and SphereFace-R v2, and they represent different types of margins as shown in Table~\ref{framework_case}. Note that SphereFace with NFN is the same as the original SphereFace~\cite{liu2017sphereface} (with additional CGD). Our methods are equivalent to the standard softmax cross-entropy loss when $m=1.0$. We vary $m$ from 1.1 to 1.5 and report the corresponding AUC-0.0005 in Table~\ref{tab:nfn}. It can be observed that small margin (\eg, 1.1) results in inferior performance, because the learned features are not sufficiently discriminative. On the other hand, incorporating a margin that is too large can not produce good results as well, due to the increased difficulty in optimization. In our experiments, $m=1.2$ achieves the best trade-off between feature discriminativeness and optimization difficulty, leading to the best performance for all types of margins.

\begin{table*}[!t]
    \centering
	\setlength{\tabcolsep}{3.57pt}
	\renewcommand{\arraystretch}{1.35}
	\caption{Evaluation on MegaFace, IJB-B and IJB-C. We use SFNet-20 as the backbone architecture and VGGFace2 as the training set for all the compared methods. Results are in $\%$ and higher number indicates better performance.} 
	\vspace{-2.5mm}
	{
	\begin{tabular}{lcccc|cc|ccc|ccc|ccc|ccc}
	& & & & & \multicolumn{2}{c|}{ MegaFace} & \multicolumn{6}{c|}{IJB-B} & \multicolumn{6}{c}{IJB-C} \\
	 & & & & & \multicolumn{2}{c|}{(refined)} & \multicolumn{3}{c|}{\scriptsize 1:1 Veri. TAR @ FAR} & \multicolumn{3}{c|}{\scriptsize 1:N Iden. TPIR @ FPIR} & \multicolumn{3}{c|}{\scriptsize 1:1 Veri. TAR @ FAR} & \multicolumn{3}{c}{\scriptsize 1:N Iden. TPIR @ FPIR} \\
	Method & FN & $m$ & $s$ & $t$ & Iden. & Veri. & 1e-6 & 1e-5 & 1e-4 & top 1 & 1e-2 & 1e-1 & 1e-6 & 1e-5 & 1e-4 & top 1 & 1e-2 & 1e-1 \\
 	\shline
 	NormFace~\cite{wang2017normface} & HFN & 0.35 & 40 & - & 76.81 & 82.18 & 32.53 & 68.20 & 82.24 & 91.17 & 58.85 & 78.99 & 65.64 & 76.31 & 86.15 & 92.09 & 70.60 & 81.43 \\
	CosFace~\cite{wang2018cosface,wang2018additive} & HFN & 0.35 & 40 & - & 81.38 & 85.73 & \textbf{40.77} & 73.66 & 85.51 & 91.96 & 67.97 & 82.77 & 70.43 & 80.21 & 88.75 & 93.09 & 75.36 & 84.90 \\
    ArcFace~\cite{deng2019arcface} & HFN & 0.4 & 40 & - & 83.80 & 87.98 & 40.15 & \textbf{76.52} & \textbf{87.50} & \textbf{92.26} & \textbf{70.25} & \textbf{85.02} & \textbf{74.32} & \textbf{82.49} & 90.17 & \textbf{93.79} & \textbf{78.22} & \textbf{86.71} \\
    Circle Loss~\cite{sun2020circle} & HFN & 0.25 & 256 & - & 83.00 & 86.28 & 36.56 & 72.81 & 86.51 & 91.41 & 65.58 & 83.73 & 69.69 & 80.66 & 89.67 & 92.96 & 75.41 & 85.63 \\
    CurricularFace~\cite{huang2020curricularface} & HFN & 0.25 & 40 & - & \textbf{88.32} & \textbf{91.82} & 22.16 & 63.35 & 88.23 & 92.66 & 47.59 & 84.93 & 35.54 & 76.49 & \textbf{91.10} & 93.73 & 54.13 & 85.77 \\
 	\hline
 	\rowcolor{Gray}
	SphereFace~\cite{liu2017sphereface} & NFN & 1.2 & - & - & 85.55 & 90.03 & \textbf{40.52} & \textbf{74.89} & \textbf{86.81} & 92.86 & \textbf{67.79} & \textbf{84.19} & \textbf{71.95} & 81.46 & \textbf{89.69} & \textbf{94.20} & 76.20 & 85.85 \\
	\rowcolor{Gray}
	SphereFace-R v1 & NFN & 1.2 & - & - & \textbf{85.63} & \textbf{90.70} & 35.42 & 73.87 & 86.59 & \textbf{92.92} & 66.75 & 83.95 & 71.60 & \textbf{81.80} & 89.67 & 94.12 & \textbf{76.29} & \textbf{85.92} \\
	\rowcolor{Gray}
	SphereFace-R v2 & NFN & 1.2 & - & - & 85.02 & 89.24 & 39.17 & 73.80 & 86.36 & 92.85 & 67.60 & 83.66 & 70.96 & 80.61 & 89.32 & 93.95 & 75.30 & 85.04 \\
	\hline
	\rowcolor{Gray}
	SphereFace & HFN & 1.2 & 30 & - & \textbf{85.44} & \textbf{90.12} & \textbf{40.11} & 75.44 & 87.43 & \textbf{92.97} & 67.70 & 84.87 & 73.79 & 83.02 & 90.37 & \textbf{94.19} & 78.18 & 86.90 \\
	\rowcolor{Gray}
	SphereFace-R v1 & HFN & 1.5 & 40 & - & 83.39 & 87.42 & 39.39 & 75.78 & 86.82 & 92.53 & 68.92 & 84.43 & 72.39 & 82.33 & 89.83 & 93.85 & 77.96 & 86.39 \\
	\rowcolor{Gray}
	SphereFace-R v2 & HFN & 1.4 & 60 & - & 85.15 & 89.35 & 36.68 & \textbf{76.61} & \textbf{87.81} & 92.85 & \textbf{70.21} & \textbf{85.50} & \textbf{74.94} & \textbf{83.51} & \textbf{90.45} & 94.06 & \textbf{78.80} & \textbf{87.24} \\
	\hline
	\rowcolor{Gray}
	SphereFace & SFN & 1.2 & 30 & 0.1 & 86.06 & 90.59 & 39.55 & 76.00 & 87.63 & 93.14 & 68.37 & 85.21 & 74.59 & 83.34 & 90.59 & 94.27 & 78.93 & 87.12 \\
	\rowcolor{Gray}
	SphereFace-R v1 & SFN & 1.5 & 40 & 0.5 & \textbf{89.17} & \textbf{91.95} & 42.38 & \textbf{80.24} & \textbf{88.88} & \textbf{93.45} & \textbf{73.55} & \textbf{87.24} & \textbf{77.61} & \textbf{85.93} & \textbf{91.59} & \textbf{94.65} & \textbf{82.09} & \textbf{89.01} \\
	\rowcolor{Gray}
	SphereFace-R v2 & SFN & 1.4 & 60 & 0.5 & 89.10 & 91.75 & \textbf{44.20} & 78.91 & 88.86 & 93.12 & 70.84 & 87.15 & 75.91 & 85.63 & 91.38 & 94.31 & 81.04 & 88.54 
	\end{tabular}}
\label{tab:vgg2_large_dataset}
\end{table*}

\subsubsection{Hard Feature Normalization}
\label{exp_hfn}
HFN introduces an additional hyperparameter $s$ to the loss function, which controls the norm of the deep features. To show how $m$ and $s$ affect the performance, we perform a grid search by varying these two hyperparameters for all three types of margins including SphereFace, SphereFace-R v1 and SphereFace-R v2. The results are given in Table \ref{tab:hfn_ablation}.

We have several observations. First, there usually exists an optimal margin hyperparameter $m$ that leads to the best performance for each scale $s$. If $m$ is smaller than the optimal value, the performance is usually improved as $m$ increases. If $m$ is greater than the optimal value, the performance is usually decreased as $m$ increases. Second, for larger $s$, the corresponding optimal $m$ will also tend to be larger. For example, in Table~\ref{tab:hfn_ablation}(b), the optimal margin $m$ is monotonously increased from 1.1 to 1.5 for the scale $s$ ranging from 20 to 60. The performance is affected by both $s$ and $m$ in a coupled manner. The same pattern also appears in Table~\ref{tab:hfn_ablation}(a) and (c).  Third, a wide range of $s$ could lead to a satisfying performance, as long as $m$ is properly tuned. For example, in Table~\ref{tab:hfn_ablation}(a), the corresponding AUC-0.0005 remains a relatively high value ($[57.09, 61.37]$) while the scale $s$ varies from 30 to 60. We note that these observations are consistent across all types of multiplicative margin.

\vspace{0.5mm}

\noindent\textbf{Hyperparameter tuning strategy.}  With the aforementioned observations, we summarize a simple yet effective strategy to search suitable hyperparameters for HFN. First, we uniformly pick the scale parameter $s$ with a relatively large gap, \eg, 20 or 30. The approximate range of $s$ can follow the common setting in many existing approaches~\cite{liu2017sphereface, deng2019arcface}. Second, we find the optimal margin parameter $m$ for each $s$ with a uniform search. It has been shown in Table~\ref{tab:hfn_ablation} that the range of $m$ depends on the specific $s$ we use. For larger $s$, its optimal $m$ is typically larger as well. As a useful practice, we should gradually search larger $m$, as $s$ increases. Finally, we choose the scale $s$ and the margin $m$ that lead to the best performance on the validation set. We find that such a simple hyperparameter searching is generally useful and can be applied to tuning different kinds of hyperspherical FR methods in practice.

\begin{figure}[!t]
  \centering
 \includegraphics[width=3.5in]{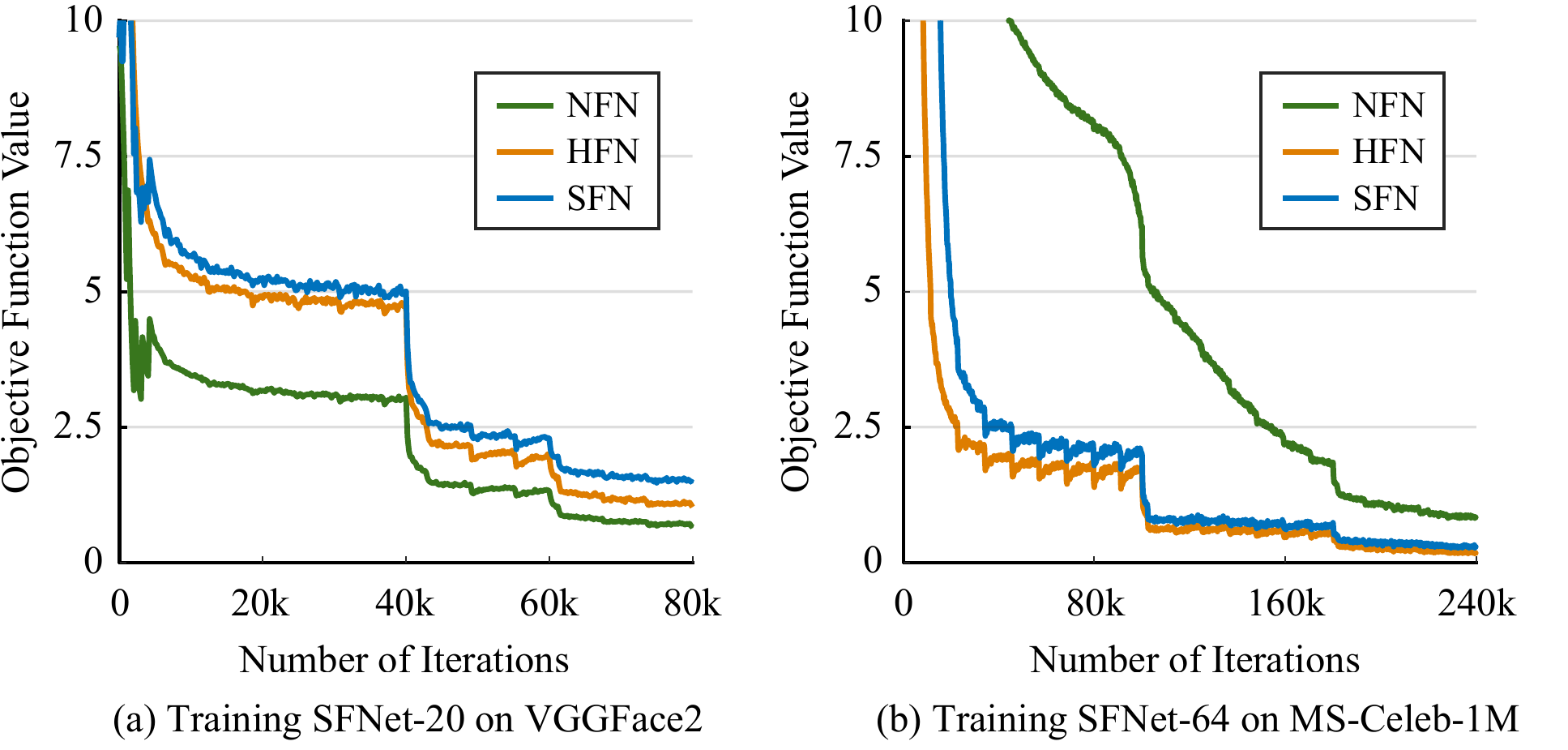}
  \vspace{-4.2mm}
  \caption{Training objective of SphereFace-R v2 with NFN, HFN and SFN on (a) VGGFace2 and (b) MS-Celeb-1M. For SFN, we only plot the softmax-based loss and the feature norm regularization is not plotted.}\label{loss_vgg2_ms1m}
\end{figure}

\subsubsection{Soft Feature Normalization}
\label{sec:sfn}
SFN is a soft regularization to constrain the feature norm and can be considered as a trade-off between NFN and HFN. SFN has a weighting hyperparameter $t$ controlling the contribution of feature norm regularization term. Since there are three hyper-parameters, \ie, $m$, $s$, and $t$, it is time-consuming and infeasible to enumerate all possible combinations of them. To efficiently evaluate SFN, we adopt the combination of $m$ and $s$ that leads to the best performance in Section~\ref{exp_hfn} and fix them throughout the SFN experiment so that we can focus on how $t$ will affect the performance.

\begin{table*}[!t]
    \centering
	\setlength{\tabcolsep}{3.53pt}
	\renewcommand{\arraystretch}{1.35}
	\caption{Evaluation on MegaFace, IJB-B and IJB-C. We use SFNet-64 as the backbone architecture and MS-Celeb-1M as the training set for all the compared methods. Results are in $\%$ and higher number indicates better performance.} 	
	\vspace{-2.5mm}
	\begin{tabular}{lcccc|cc|ccc|ccc|ccc|ccc}
	& & & & & \multicolumn{2}{c|}{MegaFace} & \multicolumn{6}{c|}{IJB-B} & \multicolumn{6}{c}{IJB-C} \\
	 & & & & & \multicolumn{2}{c|}{(refined)} & \multicolumn{3}{c|}{\scriptsize 1:1 Veri. TAR @ FAR} & \multicolumn{3}{c|}{\scriptsize 1:N Iden. TPIR @ FPIR} & \multicolumn{3}{c|}{\scriptsize 1:1 Veri. TAR @ FAR} & \multicolumn{3}{c}{\scriptsize 1:N Iden. TPIR @ FPIR} \\
	Method & FN & $m$ & $s$ & $t$ & Iden. & Veri. & 1e-6 & 1e-5 & 1e-4 & top 1 & 1e-2 & 1e-1 & 1e-6 & 1e-5 & 1e-4 & top 1 & 1e-2 & 1e-1 \\
 	\shline
	NormFace~\cite{wang2017normface} & HFN & - & 30 & - & 89.24 & 90.76 & 40.56 & 75.30 & 90.22 & 92.49 & 64.62 & 88.19 & 70.17 & 85.88 & 92.69 & 93.70 & 77.97 & 89.81 \\
    CosFace~\cite{wang2018cosface,wang2018additive} & HFN & 0.35 & 64 & - & 98.05 & 98.45 & 37.82 & 82.99 & 94.20 & 94.69 & 70.61 & 93.03 & 78.01 & 92.29 & 95.87 & 95.91 & 84.59 & 94.53 \\
    ArcFace~\cite{deng2019arcface} & HFN & 0.5 & 64 & - & \textbf{98.45} & 98.39 & 41.02 & \textbf{86.16} & \textbf{94.82} & \textbf{94.88} & \textbf{77.92} & \textbf{93.79} & \textbf{84.47} & \textbf{93.25} & \textbf{96.25} & \textbf{96.12} & \textbf{88.80} & \textbf{95.08} \\
    Circle Loss~\cite{sun2020circle} & HFN & 0.25 & 256 & - & 98.29 & \textbf{98.67} & 41.65 & 82.76 & 94.09 & 94.64 & 74.63 & 92.83 & 81.18 & 91.59 & 95.83 & 95.77 & 84.56 & 94.15 \\
    CurricularFace~\cite{huang2020curricularface} & HFN & 0.5 & 64 & - & 98.43 & 98.62 & \textbf{43.76} & 85.55 & 94.61 & 94.82 & 76.01 & 93.37 & 83.35 & 92.95 & 96.11 & 96.04 & 87.88 & 94.76 \\
 	\hline
 	\rowcolor{Gray}
	SphereFace~\cite{liu2017sphereface} & NFN & 1.1 & - & - & 93.04 & 94.37 & 37.78 & 69.99 & 88.70 & 91.98 & 60.43 & 86.28 & 62.10 & 83.03 & 91.74 & 93.30 & 71.45 & 88.11 \\
	\rowcolor{Gray}
    SphereFace-R v1 & NFN & 1.2 & - & - & 93.75 & 94.80 & 44.83 & \textbf{83.07} & 92.12 & \textbf{93.56} & 70.59 & 90.80 & \textbf{77.72} & 89.78 & 94.14 & \textbf{94.94} & 84.55 & \textbf{92.23} \\
    \rowcolor{Gray}
    SphereFace-R v2 & NFN & 1.2 & - & - & \textbf{94.74} & \textbf{95.51} & \textbf{47.82} & 82.82 & \textbf{92.15} & 93.51 & \textbf{72.38} & \textbf{90.91} & 76.14 & \textbf{89.85} & \textbf{94.17} & \textbf{94.94} & \textbf{84.58} & 92.17 \\
	\hline
	\rowcolor{Gray}
	SphereFace & HFN  & 1.7 & 32 & - & \textbf{98.16} & 98.46 & 48.83 & 86.66 & \textbf{94.36} & 94.84 & 76.35 & 93.20 & 83.57 & 92.79 & \textbf{95.82} & \textbf{96.07} & 87.74 & 94.47 \\
	\rowcolor{Gray}
	SphereFace-R v1 & HFN & 1.6 & 32 & - & 98.03 & 98.30 & \textbf{48.84} & \textbf{88.16} & 94.31 & \textbf{94.98} & \textbf{79.18} & \textbf{93.23} & \textbf{85.65} & \textbf{93.17} & 95.72 & \textbf{96.07} & \textbf{89.94} & \textbf{94.55} \\
	\rowcolor{Gray}
	SphereFace-R v2 & HFN & 1.5 & 64 & - & 98.04 & \textbf{98.48} & 45.77 & 86.52 & 94.08 & 94.70 & 74.13 & 93.13 & 82.07 & 92.61 & 95.63 & 95.92 & 88.00 & 94.32 \\
	\hline
	\rowcolor{Gray}
	SphereFace & SFN & 1.7 & 32 & 1.0 & 97.84 & \textbf{98.28} & \textbf{44.42} & 85.89 & \textbf{94.13} & \textbf{94.84} & 77.37 & 93.14 & \textbf{83.05} & 92.25 & \textbf{95.67} & 95.95 & 87.72 & 94.19 \\
	\rowcolor{Gray}
	SphereFace-R v1 & SFN & 1.6 & 32 & 1.0 & 97.42 & 97.79 & 44.24 & 84.95 & 93.70 & 94.58 & \textbf{77.66} & 92.66 & 81.92 & 91.10 & 95.27 & 95.65 & 85.70 & 93.73 \\
	\rowcolor{Gray}
	SphereFace-R v2 & SFN & 1.5 & 64 & 0.2 & \textbf{97.93} & 98.19 & 40.06 & \textbf{86.37} & 94.12 & 94.78 & 76.94 & \textbf{93.21} & 82.94 & \textbf{92.29} & 95.62 & \textbf{95.99} & \textbf{87.83} & \textbf{94.42}
	\end{tabular}
\vspace{-2mm}\label{tab:ms1m_large_dataset}
\end{table*}

We give the AUC-0.0005 in Table~\ref{tab:sfn}. SphereFace achieves 62.4\% with SFN and 61.37\% with HFN. SphereFace-R v1 achieves 61.37\% with SFN and 60.45\% with HFN. SphereFace-R v2 achieves 61.21\% with SFN and 62.72\% with HFN. Our experiments show that that SFN generally yields comparable or even better results than HFN with a properly chosen $t$ when our models are trained on VGGFace2.

\subsubsection{Characteristic Gradient Detachment}
In Table \ref{tab:CGD}, we compare the models trained with or without CGD. Except the usage of CGD, the experiments are performed with exactly the same experimental settings (\eg, dataset, architecture, training setup etc). We have evaluated CGD on all three FN strategies (\ie, NFN, HFN and SFN). The results show that CGD significantly improves the results in all scenarios, which validates the importance of simplifying the gradient of the characteristic function. CGD works particularly well for SphereFace-R v1, since it enables the gradient propagation for large margin ($m > \frac{\pi}{\theta}$ in Eq.~\ref{sfr1_eq}). Here we use the best-performing hyperparameters from our previous experiments. In fact, the consistent improvements can also be obtained from CGD with the other hyperparameters. We observe that CGD can effectively improve SphereFace and SphereFace-R under all FN strategies. Because of the consistent effectiveness of CGD, we use CGD for both SphereFace and SphereFace-R by default and the results in the other sections are also obtained with CGD.

\subsection{Ablation and Exploration on MS-Celeb-1M} \label{sec:exp_ms1m}
In this section, we conduct ablation and exploration with a deeper network that is trained on a large-scale dataset. Specifically, we use SFNet-64 as the backbone network architecture and MS-Celeb-1M as the training set. We report AUC-0.0005 on the same validation set as Section~\ref{sec:exp_vgg2}.

\subsubsection{No Feature Normalization}
We evaluate SphereFace, SphereFace-R v1 and SphereFace-R v2 without feature normalization. Note that SphereFace with NFN is equivalent to the original SphereFace~\cite{liu2017sphereface} with CGD. The results are given in Table~\ref{tab:ms1m-nfn}. Similar to the experiments on VGGFace2, NFN generally works well with small margins, achieving 79.89\%, 84.66\%, and 84.78\% with $m$ being 1.1, 1.2, and 1.2, respectively. The increased number of training identities (from 8K to 86K) leads to severe optimization difficulty during training. This can be empirically observed from the smaller optimal $m$ from SphereFace and the dramatically decreased performance from both SphereFace and SphereFace-R with larger margins.

\subsubsection{Hard Feature Normalization}
Following our hyperparameter tuning strategy given in Section~\ref{exp_hfn}, we search the optimal combination of $s$ and $m$. Since the performance is stable across a wide range of $s$, here we use a relatively large gap (\ie, 32). As mentioned in Section~\ref{exp_hfn}, the optimal $m$ tends to increase with larger $s$. We search two feature scale hyperparameters for SphereFace (\ie, $s=32,64$), SphereFace-R v1 (\ie, $s=32,64$) and SphereFace-R v2 (\ie, $s=64,128$). Based on our observation on VGGFace2, the optimal $m$ for larger $s$ also tends to be larger. Therefore for smaller $s$, we search $m$ from 1.4 to 1.8, and for larger $s$, we search $m$ from 1.6 to 2.0. The results in Table~\ref{tab:ms1m-hfn} well match our expectation that there is only one optimal $m$ for each $s$ and the optimal $m$ will increase with larger $s$. Under the same $s$, we discover that as $m$ deviates from its optimal value, the performance will also become worse. The distribution of the performance for different $m$ exhibits a strong unimodality, which can largely benefit the hyperparameter tuning. The best models of SphereFace, SphereFace-R v1 and SphereFace-R v2 achieve 91.03\%, 90.78\% and 91.58\% with $s=32$, $32$, and $64$, respectively. The performance on MS-Celeb-1M is also much better than that on VGGFace2, which shows that our methods can easily enjoy the accuracy boost from larger training set and deeper network. Moreover, the performance of both SphereFace and SphereFace-R is not very sensitive to $m$ and remains stable for a wide range of $m$.

\subsubsection{Soft Feature Normalization}
Similar to Section~\ref{sec:sfn}, we adopt the best-performing settings ($m$ and $s$) from HFN and vary the hyperparameter $t$ in order to evaluate the performance of SFN. The results are reported in Table~\ref{tab:sfn_ms1m}. With a properly tuned $t$, SFN achieves 87.70\%, 87.19\% and 86.10\% AUC-0.0005 for SphereFace, SphereFace-R v1 and SphereFace-R v2, respectively. We observe that the performance of SFN on MS-Celeb-1M are not as good as HFN, which contradicts the observation on VGGFace2. This implies that SFN may be sensitive to the distribution of the training data. We hypothesize that SFN is more sensitive to noisy samples (since MS-Celeb-1M has much more low-quality and noisy images than VGGFace2).

\begin{table*}[!t]
    \centering
	\setlength{\tabcolsep}{3.38pt}
	\renewcommand{\arraystretch}{1.35}
	\caption{{\color{modify}Evaluation on MegaFace, IJB-B and IJB-C. We use IResNet-100 as the backbone architecture and MS-Celeb-1M as the training set. Results are in $\%$ and higher number indicates better performance. For ArcFace, we present results from their paper and our re-implementation.}} 
	\vspace{-2.5mm}
	\color{modify}\begin{tabular}{lcccc|cc|ccc|ccc|ccc|ccc}
	& & & & & \multicolumn{2}{c|}{MegaFace} & \multicolumn{6}{c|}{IJB-B} & \multicolumn{6}{c}{IJB-C} \\
	 & & & & & \multicolumn{2}{c|}{(refined)} & \multicolumn{3}{c|}{\scriptsize 1:1 Veri. TAR @ FAR} & \multicolumn{3}{c|}{\scriptsize 1:N Iden. TPIR @ FPIR} & \multicolumn{3}{c|}{\scriptsize 1:1 Veri. TAR @ FAR} & \multicolumn{3}{c}{\scriptsize 1:N Iden. TPIR @ FPIR} \\
	Method & FN & $m$ & $s$ & $t$ & Iden. & Veri. & 1e-6 & 1e-5 & 1e-4 & top 1 & 1e-2 & 1e-1 & 1e-6 & 1e-5 & 1e-4 & top 1 & 1e-2 & 1e-1 \\
 	\shline
 	CosFace \cite{wang2018additive}, \cite{wang2018cosface} & HFN & 0.4 & 64 & - & \textbf{98.70} & \textbf{98.87} & \textbf{43.67} &	88.83 & \textbf{95.23} & \textbf{95.35} & 80.50 & \textbf{94.49} & 85.29 & 94.33 & \textbf{96.62} & \textbf{96.53} &	90.69 & \textbf{95.61} \\
    ArcFace (results in \cite{deng2019arcface}) & HFN & 0.5 & 64 & - & 98.35 & 98.48 & 38.28	& 89.33 & 94.25 & - & - & - & \textbf{86.25} & 93.15 & 95.65 & - & - & - \\
    ArcFace ~\cite{deng2019arcface} & HFN & 0.5 & 64 & - & 98.67 & 98.46 & 43.43 & \textbf{90.40} & 95.02 & 95.14 & \textbf{81.36} & 94.26 & 86.00 & \textbf{94.49} & 96.39 & 96.47 & \textbf{91.91} & 95.51 \\
 	\hline
 	\rowcolor{Gray}
	SphereFace~\cite{liu2017sphereface} & NFN & 1.1 & - & - & 94.95 & 95.76 & \textbf{43.02} & 73.79 & 90.19 & 92.67 & 64.09 & 87.80 & 68.83 & 85.77 & 92.82 & 93.89 & 76.83 & 89.93 \\
	\rowcolor{Gray}
    SphereFace-R v1 & NFN & 1.2 & - & - & 96.49 & 97.22 & 38.01 & 81.31 & 92.58 & 94.15 & 70.49 & 91.20 & 77.94 & 90.23 & 94.83 & 95.34 & \textbf{84.81} & 92.91 \\
    \rowcolor{Gray}
    SphereFace-R v2 & NFN & 1.2 & - & - & \textbf{97.14} & \textbf{97.56} & 41.73 & \textbf{82.60} & \textbf{93.00} & \textbf{94.42} & \textbf{70.93} & \textbf{91.44} & \textbf{78.64} & \textbf{90.70} & \textbf{95.03} & \textbf{95.45} & 84.75 & \textbf{93.11} \\
	\hline
	\rowcolor{Gray}
	SphereFace & HFN  & 1.7 & 32 & - & \textbf{98.63} & \textbf{99.04} & 47.33 & \textbf{90.14} & 94.87 & 95.13 & 82.57 & \textbf{94.30} & \textbf{87.86} & \textbf{94.36} & \textbf{96.25} & \textbf{96.45} & \textbf{91.68} & \textbf{95.36} \\
	\rowcolor{Gray}
	SphereFace-R v1 & HFN & 1.6 & 32 & - & 98.61 & 98.61 & \textbf{47.82} & 89.06 & \textbf{94.89} & \textbf{95.18} & \textbf{82.69} & 94.09 & 86.30 & 93.91 & 96.21 & 96.38 & 91.16 & 95.19 \\
	\rowcolor{Gray}
	SphereFace-R v2 & HFN & 1.5 & 64 & - & 98.46 & 98.69 & 41.22 & 88.86 & 94.77 & 95.18 & 79.57 & 93.78 & 84.21 & 93.55 & 96.14 & 96.27 & 89.85 & 94.95 \\
	\hline
	\rowcolor{Gray}
	SphereFace & SFN & 1.7 & 32 & 1.0 & 98.23 & 98.50 & 45.35 & 84.86 & 94.20 & 94.90 & 76.38 & 93.32 & 80.83 & 92.17 & 95.70 & 95.91 & 86.44 & 94.37 \\
	\rowcolor{Gray}
	SphereFace-R v1 & SFN & 1.6 & 32 & 1.0 & 98.07 & 98.15 & 42.97 & \textbf{87.92} & 94.32 & 94.94 & \textbf{78.47} & 93.32 & \textbf{85.43} & \textbf{93.06} & 95.83 & 96.02 & \textbf{89.13} & 94.56 \\
	\rowcolor{Gray}
	SphereFace-R v2 & SFN & 1.5 & 64 & 0.2 & \textbf{98.26} & \textbf{98.58} & \textbf{45.64} & 86.55 & \textbf{94.51} & \textbf{95.10} & 76.53 & \textbf{93.65} & 80.19 & 93.01 & \textbf{95.96} & \textbf{96.19} & 87.39 & \textbf{94.88} 
	\end{tabular}
\vspace{-2mm}\label{tab:iresnet100}
\end{table*}

\subsection{Evaluation on Large-scale Benchmarks}\label{exp_benchmark}
\subsubsection{Experiments with SFNet-20 and SFNet-64}
In this section, we evaluate the performance of both SphereFace and SphereFace-R with three FN strategies on popular large-scale benchmarks (\ie, IJB-B, IJB-C and MegaFace). We also provide fair comparisons to state-of-the-art methods (with the same training set and backbone network). We use the same models from Section~\ref{sec:exp_vgg2} and Section~\ref{sec:exp_ms1m} with the best-performing hyperparameters on the validation set. Specifically, we train SFNet-20 on VGGFace2 and SFNet-64 on MS-Celeb-1M. We compare our models to current state-of-the-art methods, \ie, NormFace~\cite{wang2017normface}, CosFace~\cite{wang2018cosface,wang2018additive}, ArcFace~\cite{deng2019arcface}, circle loss~\cite{sun2020circle} and CurricularFace~\cite{huang2020curricularface}. The hyperparameters of these methods are tuned to achieve to the best validation performance.

We make several useful observations from Table~\ref{tab:vgg2_large_dataset} and Table~\ref{tab:ms1m_large_dataset}. First, the results on large-scale testing sets are consistent with those on the validation set, especially the metrics at low false acceptance rate (FAR) or false positive identification rates (FPIR). The consistent performance demonstrates the effectiveness of the selected validation set and metric. This indicates that the models that achieve higher performance on the validation set usually show better results on MegaFace and IJB as well. 

Second, different types of margins tend to achieve similar performance, while different types of FN strategies have significantly different results. In Fig.~\ref{loss_vgg2_ms1m}, we first show the training losses for different FN strategies. Fig.~\ref{loss_vgg2_ms1m}(a) shows that all the models equipped with NFN, HFN or SFN converge well on VGGFace2, a relatively small, clean and high-quality training set. From Table~\ref{tab:vgg2_large_dataset}, NFN and HFN show comparable generalization ability, while SFN achieves the best performance among all FN strategies. This implies that making good use of the magnitude information during training can effectively improve the results. Fig.~\ref{loss_vgg2_ms1m}(b) shows that NFN is unable to converge to a sufficiently small training loss on MS-Celeb-1M, a large, noisy and low-quality dataset. From Table~\ref{tab:ms1m_large_dataset}, we can observe that NFN indeed converges to a bad local minima that generalizes poorly. By introducing a magnitude regularization term to the objective function, SFN can effectively help the models escaping from the bad local minima. In contrast to the results on VGGFace2, we find that SFN performs worse than HFN on MS-Celeb-1M, implying that SFN may be more sensitive to noisy samples in the training set and HFN may be more robust to different training sets than NFN and SFN.

Finally, with our proposed modifications, all variants in our SphereFace family shows competitive results compared to the state-of-the-art methods.  Both SphereFace and SphereFace-R perform particularly well under the low FAR, such as 1:1 verification TAR at 1e-6 and 1:N identification TPIR at 1e-1 FPIR. These metrics are very important in designing a robust face recognition system in practice.

\subsubsection{{\color{modify}Experiments with IResNet-100}}\label{IResNet100}
{\color{modify}In order to have a comprehensive comparison with the published results, we conduct the experiments to train our methods on MS-Celeb-1M~\cite{guo2016ms} with IResNet-100~\cite{deng2019arcface}. Comparing the results in Table~\ref{tab:iresnet100} and Table~\ref{tab:ms1m_large_dataset}, we observe that IResNet100 achieves better performance than those using SFNet-64 (with the same set of hyperparameters), establishing a higher baseline for SphereFace. For different FN strategies, IResNet100 performs similarly to SFNet64 in the sense that HFN is slightly better than SFN and they are both better than NFN. In general, both SphereFace and SphereFace-R are generally comparable to CosFace and ArcFace. SphereFace with HFN achieves the best performance on MegaFace. More interestingly, we find that both SphereFace and SphereFace-R v1 with HFN achieve significantly better 1:1 verification performance at low FAR than all the compared methods on IJB.}

\section{Concluding Remarks}
Our paper proposes a novel framework that unifies hyperspherical face recognition. This framework provides a general principle for a loss function to incorporate large angular margins. Under this framework, we substantially extend and improve our previous work on SphereFace~\cite{liu2017sphereface} by addressing training instability and significantly improving empirical performance. Specifically, we propose two new types of multiplicative margins that effectively implement the original intuition of SphereFace. Moreover, we also come up with a novel implementation technique called characteristic gradient detachment to further improve training stability and generalization. Extensive experiments on a number of popular benchmarks are conducted to validate the superiority of our SphereFace family.

Based on the unified framework, our paper demonstrates strong flexibility and many unique advantages of hyperspherical FR. There still exist a number of exciting yet under-explored open problems in hyperspherical FR, such as how to design better angular margin, how to effectively incorporate feature magnitude into training and testing, how to learn the loss function directly from data, etc. We also present a few useful characterizations for the loss function in hyperspherical FR, leading to multiple equivalent loss design spaces. Current popular loss functions only represent a very small and limited 
subset in the 
huge design space of hyperspherical FR. We expect that more 
work can be devoted to this promising line of research in the future.


\ifCLASSOPTIONcompsoc
  \section*{Acknowledgments}
\else
  \section*{Acknowledgment}
\fi

The authors would like to sincerely thank Haoran Sun, Yuyu Zhang and Will Powell for generously helping us to schedule computing resources, and Hanchen Wang for proofreading. Weiyang Liu is supported by a Cambridge-Tübingen Fellowship, an NVIDIA GPU grant, DeepMind and the Leverhulme Trust via CFI. Adrian Weller acknowledges support from a Turing AI Fellowship under grant EP/V025379/1, The Alan Turing Institute under EPSRC grant EP/N510129/1 and TU/B/000074, and the Leverhulme Trust via CFI. This work is partially supported by the Defence Science and Technology Agency (DSTA), Singapore under contract number A025959, and this paper does not reflect the position or policy of DSTA and no official endorsement should be inferred.

{
\bibliographystyle{IEEEtran}
\bibliography{egbib}
}

\ifCLASSOPTIONcaptionsoff
  \newpage
\fi

\vspace{-7mm}
\begin{IEEEbiography}[{\includegraphics[width=1in,height=1.25in,clip,keepaspectratio]{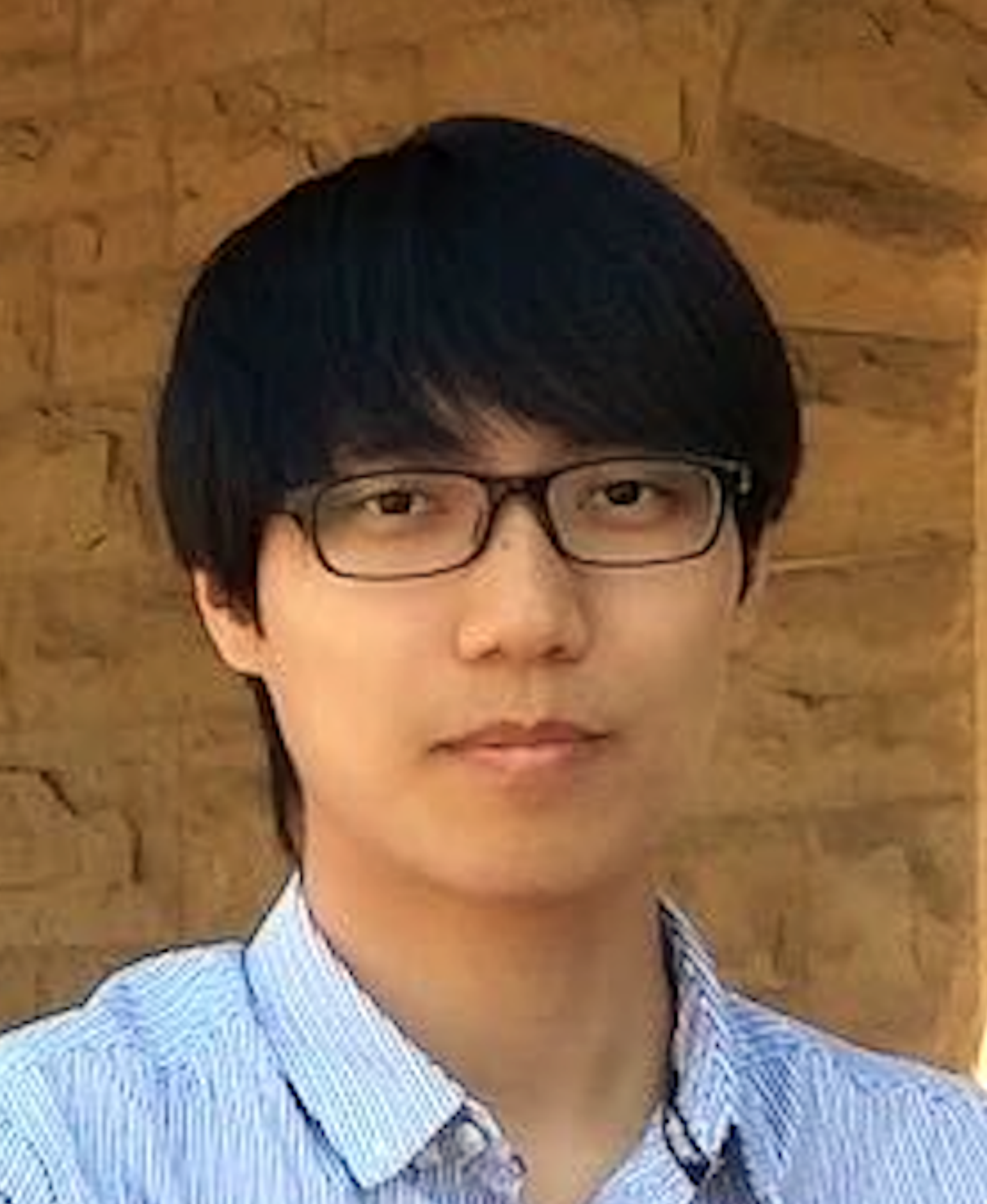}}]{Weiyang Liu} 
is currently conducting research at the University of Cambridge, UK and the Max Planck Institute for Intelligent Systems, Tübingen, Germany under the Cambridge-Tübingen Fellowship Program. Prior to joining this program, he has been with College of Computing, Georgia Institute of Technology, Atlanta, GA, USA. His research interests broadly lie in deep learning, representation learning, interactive machine learning and causality.
\end{IEEEbiography}

\vspace{-7mm}

\begin{IEEEbiography}[{\includegraphics[width=1in,height=1.25in,clip,keepaspectratio]{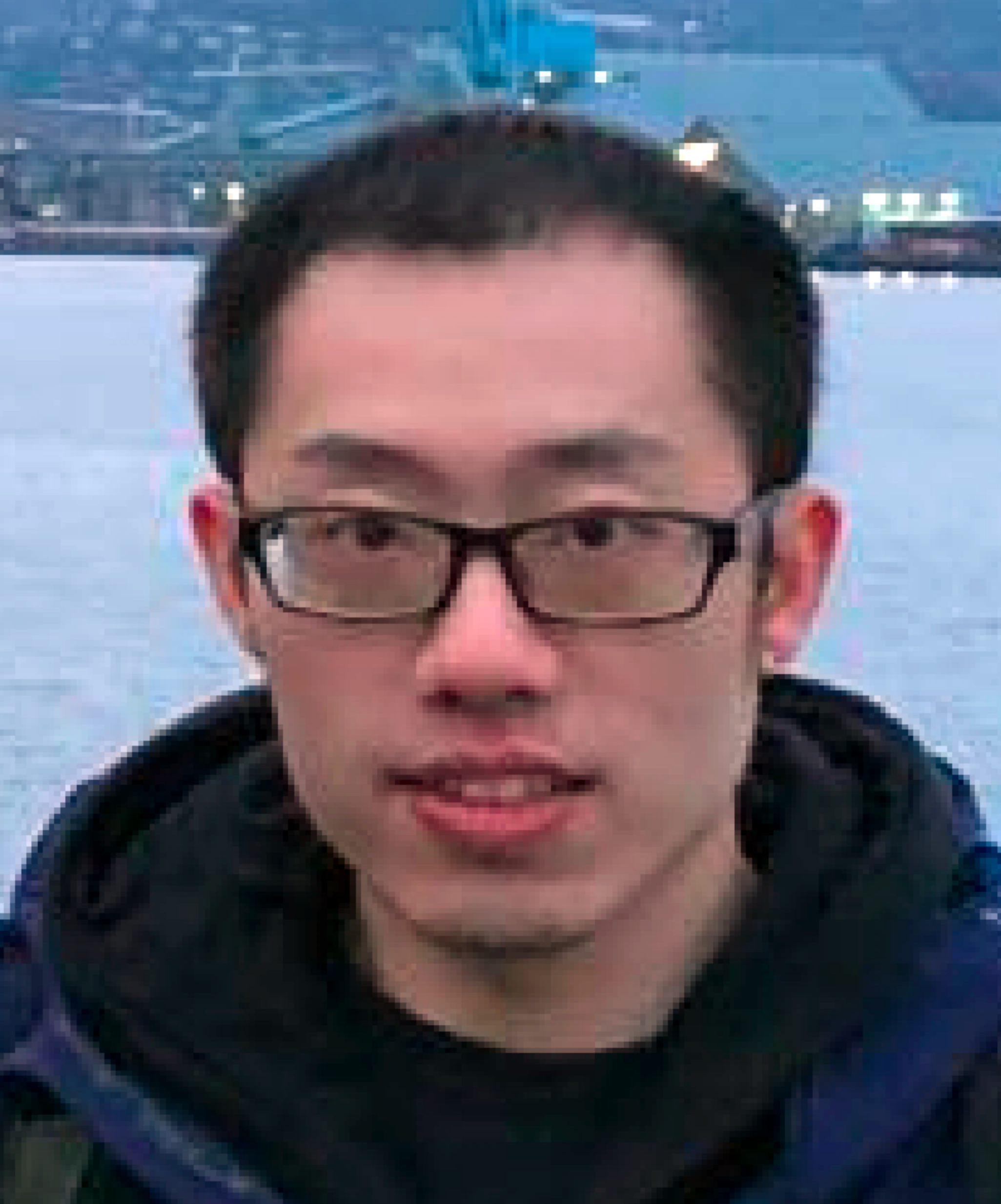}}]{Yandong Wen}
received the B.S. and M.S. degrees from South China University of Technology, Guangzhou, China in 2013 and 2016, respectively. He is a Ph.D. candidate at Carnegie Mellon University, Pittsburgh, PA, USA, where he works with Bhiksha Raj and Rita Singh. His current research interests are deep learning for face recognition, audio-visual association learning, and 3D face reconstruction.
\end{IEEEbiography}

\vspace{-7mm}


\begin{IEEEbiography}[{\includegraphics[width=1in,height=1.25in,clip,keepaspectratio]{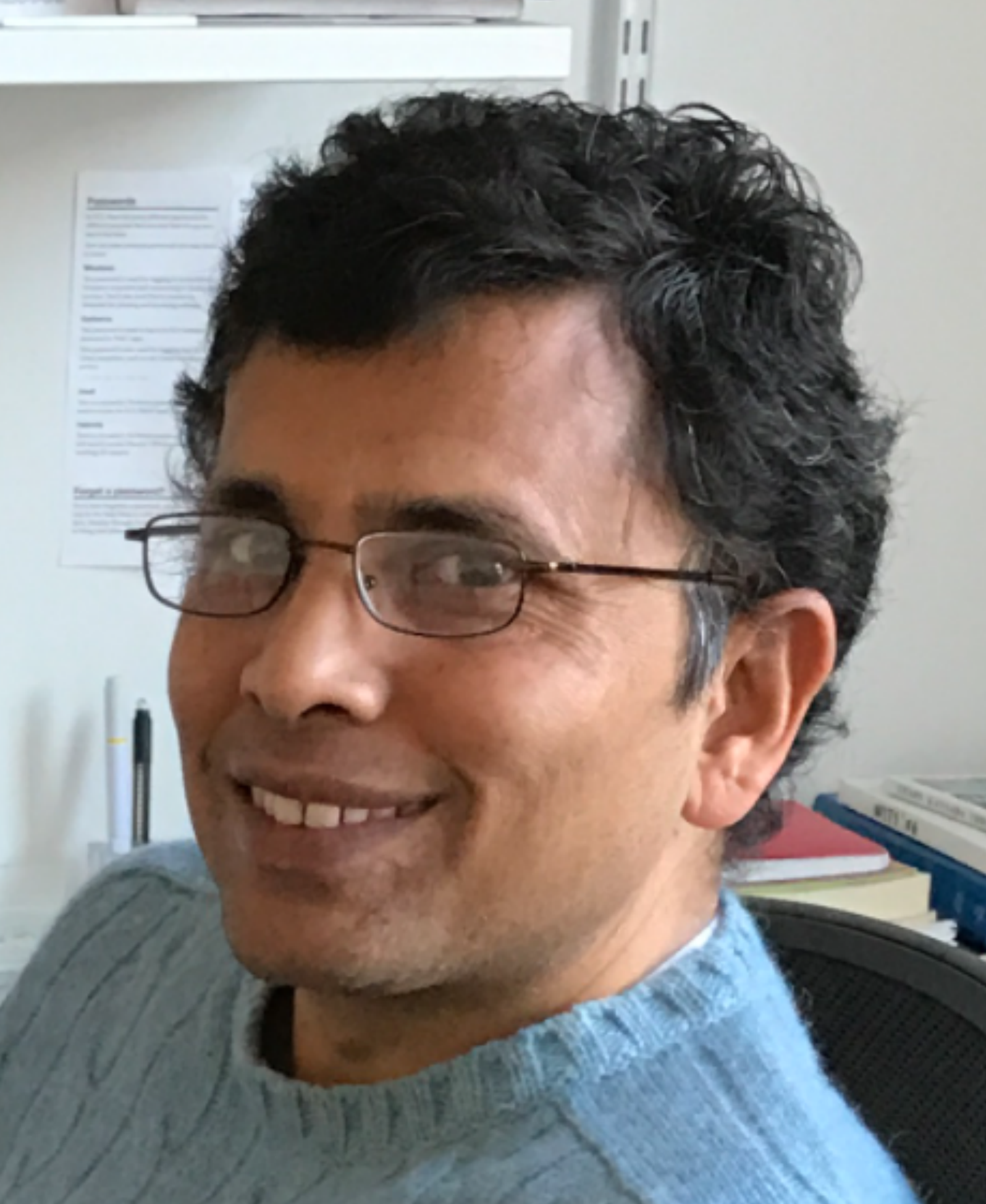}}]{Bhiksha Raj}(Fellow, IEEE)
received the Ph.D. degree in electrical and computer engineering from Carnegie Mellon University, Pittsburgh, PA, USA, in 2000. He is a Professor of the Computer Science Department, Carnegie Mellon University where he leads the Machine Learning for Signal Processing Group. He joined the Carnegie Mellon faculty in 2009, after spending time at the Compaq Cambridge Research Labs and Mitsubishi Electric Research Labs. He has devoted his career to developing speech- and audio-processing technology. He has had several seminal contributions in the areas of robust speech recognition, audio content analysis and signal enhancement, and has pioneered the area of privacy-preserving speech processing. He is also the Chief Architect of the popular Sphinx-4 speech-recognition system.
\end{IEEEbiography}



\begin{IEEEbiography}[{\includegraphics[width=1in,height=1.25in,clip,keepaspectratio]{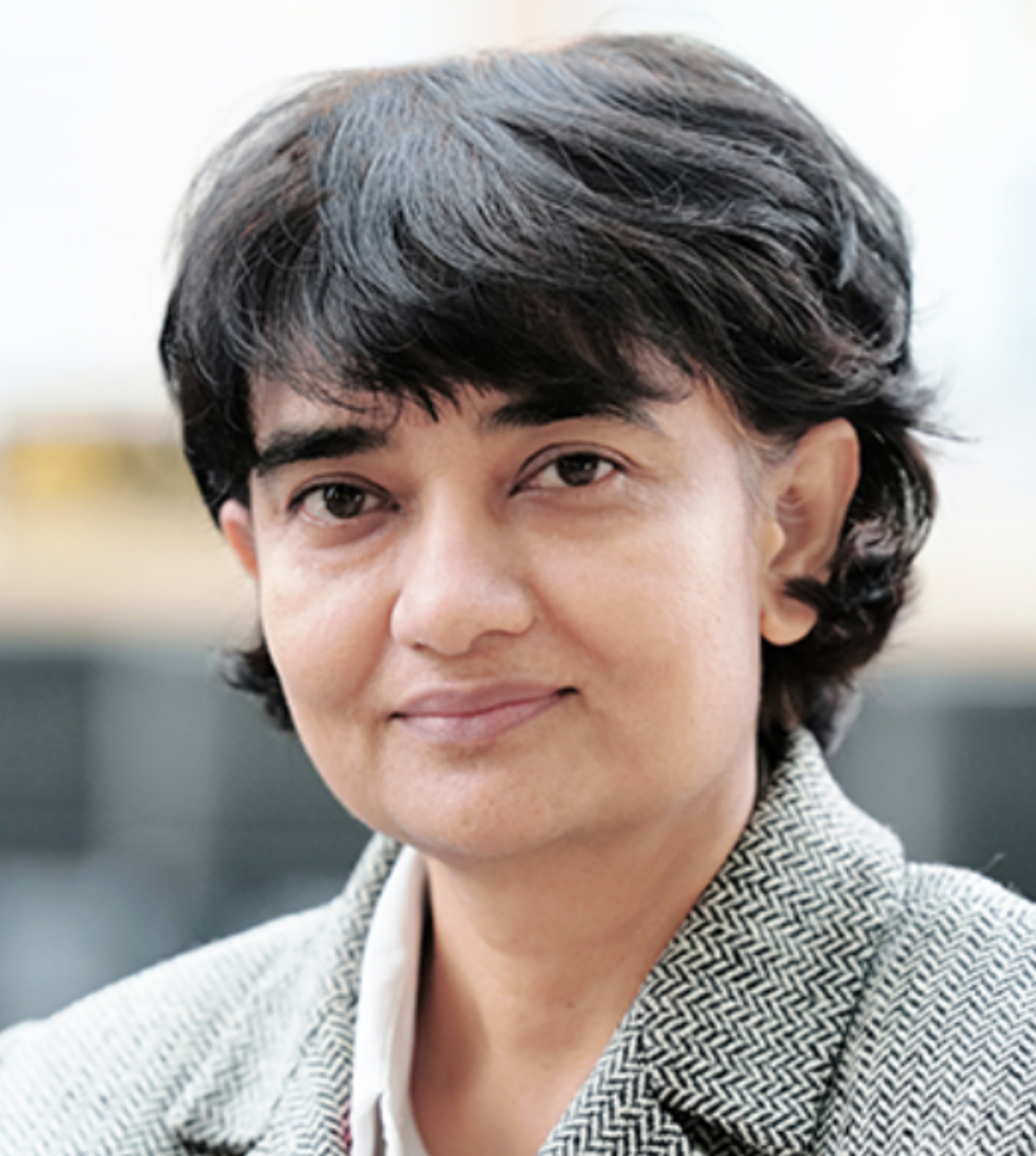}}]{Rita Singh}
received the B.Sc.(Hons.) degree in physics and the M.Sc. degree in exploration geophysics, both from the Banaras Hindu University, India. She received the Ph.D degree in geophysics in 1996 from the National Geophysical Research Institute of the Council of Scientific and Industrial Research, India. She is currently a Member of the Research Faculty at the School of Computer Science, Carnegie Mellon University (CMU), Pittsburgh, PA, USA. 
From March 1996 to November 1997, she was a Postdoctoral Fellow with the Tata Institute of Fundamental Research, India, where she worked with the Condensed Matter Physics and Computer Systems and Communications Groups. During this period, she worked on nonlinear dynamical systems and signal processing as an extension of her doctoral work on nonlinear geodynamics and chaos. 
Since November 1997, she has been affiliated with the Robust Speech Recognition and SPHINX Groups at CMU. She currently works on core algorithmic aspects of computer voice recognition, and artificial intelligence applied to voice forensics. Her focus is on the development of technology for the automated discovery, measurement, representation and learning of the information encoded in voice signal for optimal voice intelligence.
\end{IEEEbiography}

\begin{IEEEbiography}[{\includegraphics[width=1in,height=1.25in,clip,keepaspectratio]{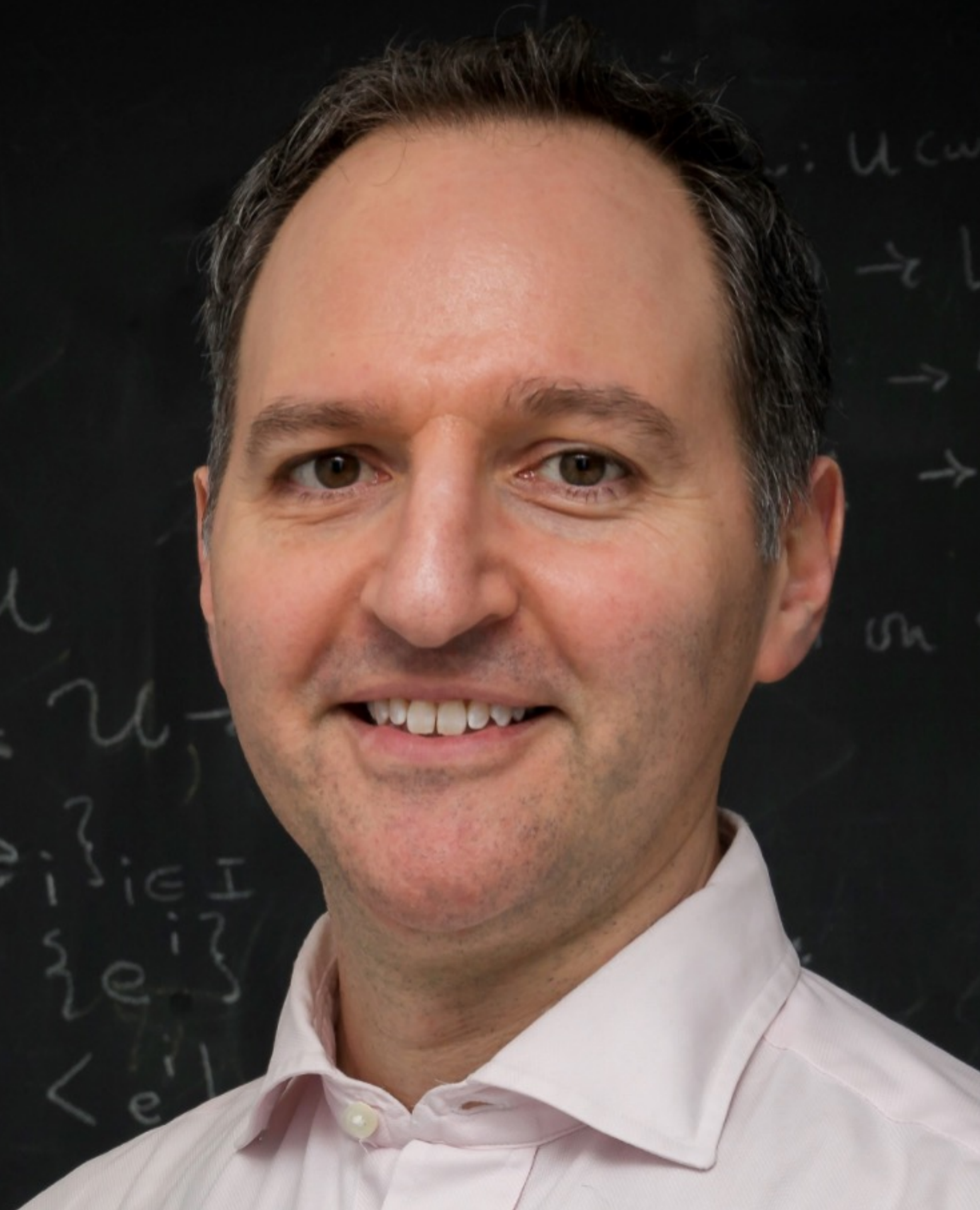}}]{Adrian Weller}
received his undergraduate degree from the University of Cambridge, and his PhD from Columbia University in New York. He is Programme Director for AI at The Alan Turing Institute, the UK national institute for data science and AI, where he is also a Turing Fellow leading work on safe and ethical AI. He is a Principal Research Fellow in Machine Learning at Cambridge, and at the Leverhulme Centre for the Future of Intelligence where he is Programme Director for Trust and Society. His interests span AI, its commercial applications and helping to ensure beneficial outcomes for society. He is Co-Director of the European Laboratory for Learning and Intelligent Systems (ELLIS) programme on Human-centric Machine Learning. Previously, he held senior roles in finance.
\end{IEEEbiography}
\vfill

\end{document}